\documentclass[10pt,twocolumn,letterpaper]{article}

\usepackage[pagenumbers]{cvpr} 

\usepackage{booktabs}
\usepackage{multirow}
\usepackage{rotating}
\usepackage{adjustbox}
\usepackage{bm}
\usepackage{soul}
\def\eg{\emph{e.g}\onedot}

\def\ie{\emph{i.e}\onedot}

\newcommand{\myparagraph}[1]{\vspace{4pt}\noindent\textbf{#1}}

\definecolor{cvprblue}{rgb}{0.21,0.49,0.74}
\usepackage[pagebackref,breaklinks,colorlinks,allcolors=cvprblue]{hyperref}

\def\confName{CVPR}
\def\confYear{2025}

\title{To Match or Not to Match: Revisiting Image Matching\\for Reliable Visual Place Recognition}

\author{Davide Sferrazza$^{1,2}$ \quad Gabriele Berton$^{2}$ \quad Gabriele Trivigno$^{2}$
\quad
Carlo Masone$^{2,3}$\\
$^{1,2}$ Politecnico di Torino\quad
$^{3}$ Focoos AI \\
{\tt\small $^{1}\,$\{name.surname\}@studenti.polito.it \, $^{2}\,$\{name.surname\}@polito.it \, $^{3}\,$\{name.surname\}@focoos.ai}
}

\begin{document}
\maketitle
\begin{abstract}
Visual Place Recognition (VPR) is a critical task in computer vision, traditionally enhanced by re-ranking retrieval results with image matching. However, recent advancements in VPR methods have significantly improved performance, challenging the necessity of re-ranking. In this work, we show that modern retrieval systems often reach a point where re-ranking can degrade results, as current VPR datasets are largely saturated. We propose using image matching as a verification step to assess retrieval confidence, demonstrating that inlier counts can reliably predict when re-ranking is beneficial. Our findings shift the paradigm of retrieval pipelines, offering insights for more robust and adaptive VPR systems. The code is available at {\small\url{https://github.com/FarInHeight/To-Match-or-Not-to-Match}}.
\end{abstract}    

\section{Introduction}
\label{sec:introduction}

Visual Place Recognition (VPR) addresses the fundamental question: “Where was this picture taken?”.
VPR is typically framed as an image retrieval problem, where a query image is localized by comparing it to a database of geotagged images \cite{Zaffar_2021_vprbench, Arandjelovic_2018_netvlad, Kim_2017_crn, Ge_2020_sfrs, Berton_2022_cosPlace, Alibey_2023_mixvpr, Alibey_2022_gsvcities, Zhu_2023_r2former, Leyvavallina_2021_gcl}. 
and it serves as a critical first step in applications such as Structure-from-Motion
(SfM) \cite{Li_2018_megadepth, schoenberger2016sfm, Liu_2017_sfm}, simultaneous localization and mapping (SLAM) \cite{Sarlin_2020_superglue, Hausler_2021_patch_netvlad, cummins2008fab} and Visual Localization \cite{Li_2018_megadepth, Sattler_2018_aachen_daynight, trivigno2024unreasonable, Torii_2018_tokyo247}.
To address this task in large-scale environments, a comprehensive database is required, which is often composed of daytime Street View images \cite{Torii_2015_pitts250k, Torii_2018_tokyo247, Berton_2021_svox, Berton_2022_cosPlace}.
However,  real-world queries may exhibit significant appearance variations due to nighttime conditions, occlusions, or adverse weather.
This domain shift between queries and database images remains a major obstacle in VPR research \cite{Torii_2018_tokyo247, Asha_2019_todaygan, Wang_2019_DA_for_VPR, Berton_2021_svox, Warburg_2020_msls, Ibrahimi_2021_insideout_vpr, Yildiz_2022_AmsterTime, Zhang_2021_survey}.
Hence, a common strategy to improve performance in VPR systems is to adopt a post-processing step to refine retrieval predictions \cite{Barbarani_2023_local, Wang_2022_TransVPR, Hausler_2021_patch_netvlad}; the underlying idea being that one can apply a more computationally-intensive method on a shortlist of candidate to filter out outliers, which would be too expensive and time-consuming to apply to the entire database.
Given the large corpus of literature on re-ranking \cite{Noh_2017_delf, Cao_2020_delg, Lee_2022_cvnet, Wang_2022_TransVPR, Hausler_2021_patch_netvlad} and image matching \cite{Revaud_2019_r2d2, Dusmanu_2019_D2Net, Sun_2021_loftr, Sarlin_2020_superglue}, this two-step pipeline established itself as the de-facto standard to refine retrieval predictions. As local features are inherently more robust to domain shifts, occlusions and perspective changes, it has been repeatedly shown that this strategy can lead to large improvements in results \cite{Hausler_2021_patch_netvlad, Wang_2022_TransVPR, Barbarani_2023_local}.

\begin{figure}[t]
    \begin{center}
    \includegraphics[width=\linewidth]{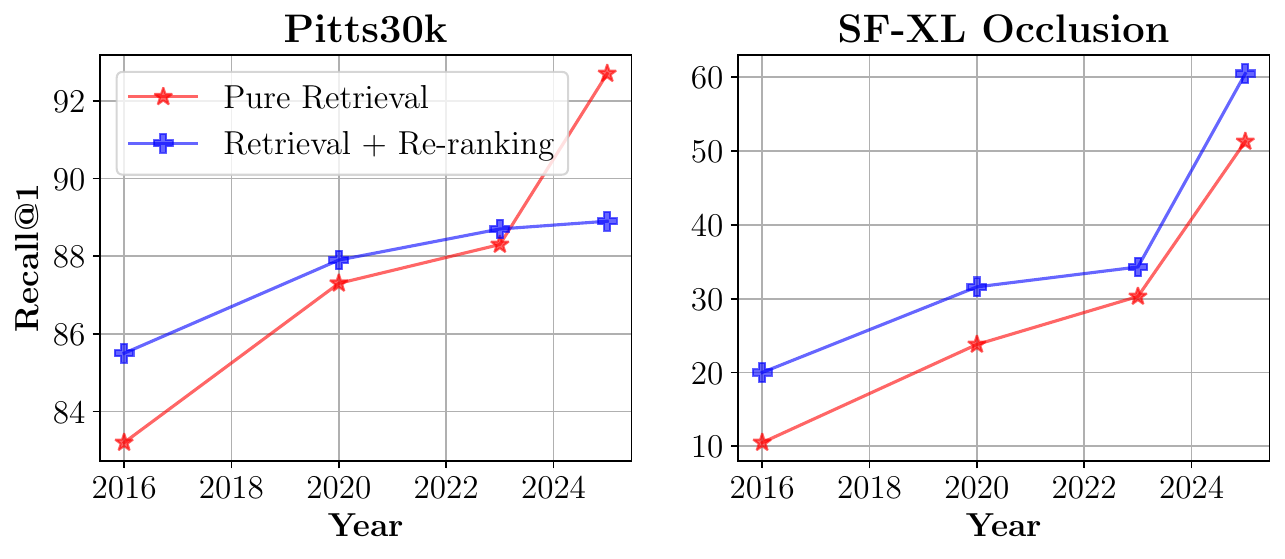}
    \end{center}
    \vspace{-5mm}
    \caption{\textbf{Re-ranking with SuperGlue with VPR methods from different years} (NetVLAD~\cite{Arandjelovic_2018_netvlad}, SFRS~\cite{Ge_2020_sfrs}, EigenPlaces~\cite{Berton2023-eigenplaces}, MegaLoc~\cite{Berton_2025_megaloc}).
    In the past, re-ranking the top-$K$ VPR results with powerful image matching methods was guaranteed to improve results.
    With modern VPR models, this is now true only for certain datasets or types of images.
    This paper explores this phenomenon, aiming to determine whether re-ranking can be adaptively and confidently triggered for individual queries during deployment.
    }
    \label{fig:teaser}
\vspace{-3mm}
\end{figure}

\begin{figure}[t]
\centering
\includegraphics[width=0.99\linewidth]{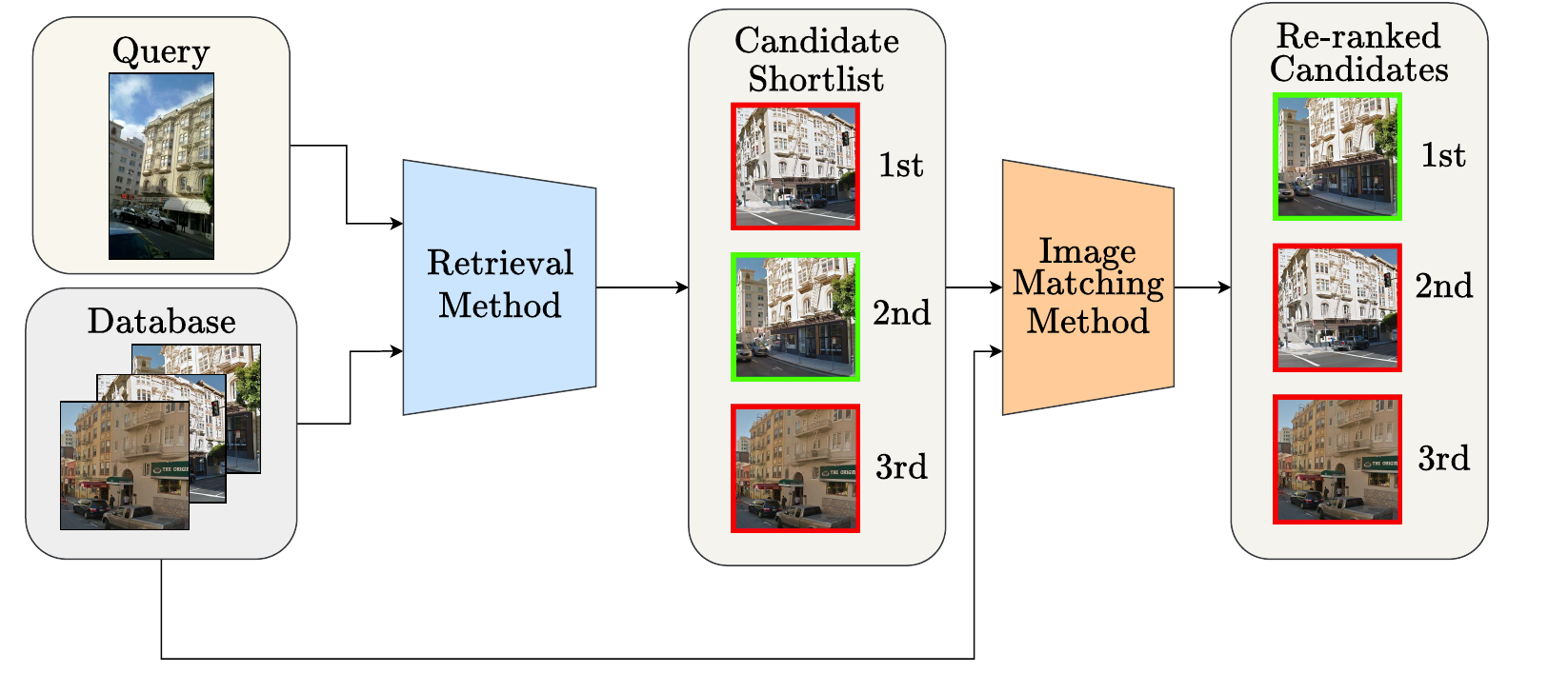}
\caption{\textbf{Re-ranking pipeline.} The standard re-ranking pipeline consists of first retrieving a shortlist of candidates using a retrieval method, followed by sorting these candidates in descending order based on the number of inliers computed using an image matching method.}
\vspace{-3mm}
\label{fig:diagram}
\end{figure}

Recent advances in VPR literature, such as the introduction of methods based on DINOv2 \cite{oquab2023-dinov2}, combined with task-specific aggregations and mining techniques \cite{Izquierdo2024-clique, Berton_2025_megaloc, Izquierdo2024-salad} achieved unprecedented results, showing remarkable generalization capabilities.
In this work, we propose a \textit{reality check} on the performance of modern VPR and image matching methods, showing that recent advancements have caused a paradigm shift in the typical retrieval+re-ranking pipeline. Specifically, we show in \cref{fig:teaser} that (i) modern retrieval methods have reached the point where applying re-ranking can, surprisingly, worsen performance in some cases; and that (ii) current VPR datasets are largely saturated by the current state-of-the-art.  
The main takeaway from the preliminary experiments in \cref{fig:teaser} is that applying a re-ranking step (\textit{cf}. \cref{fig:diagram}) is not always beneficial. This raises the question of whether an automated approach can discern when retrieval predictions already possess sufficient confidence, preventing potentially detrimental post-processing.

In this work we demonstrate that the number of inliers can serve as a proxy of prediction uncertainty, in turn providing an indication of whether a re-ranking step can improve retrieval performance or not.
In essence, we argue that image matching methods should be employed first as a \textit{verification} step, to assess the confidence of the retrieval predictions, and only afterwards, it should be selectively used as a \textit{postprocessing} step for the uncertain estimates. This finding derives from a comprehensive evaluation of image matching methods in both roles, which sets this study apart from prior works that focus solely on re-ranking performance \cite{Hausler_2021_patch_netvlad, Wang_2022_TransVPR, Barbarani_2023_local}. 

\begin{figure}
\centering

\begin{subfigure}{0.47\linewidth}
  \centering
  \includegraphics[width=\linewidth]{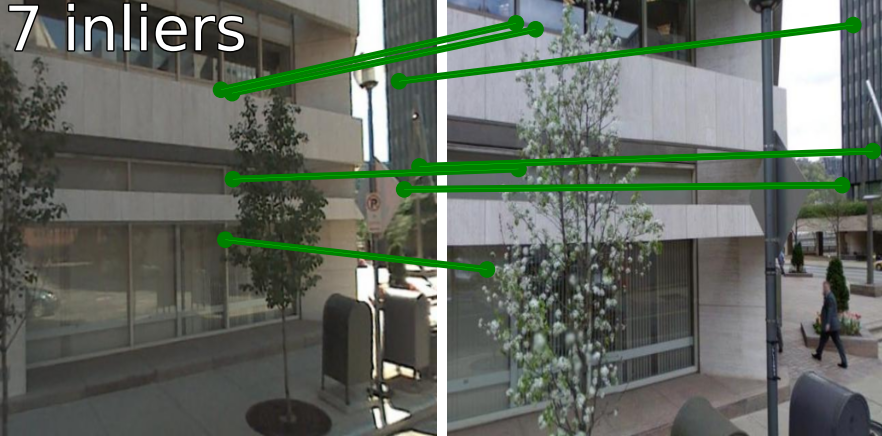}
  \caption{A query and its positive}
\end{subfigure}%
\hfill
\begin{subfigure}{0.47\linewidth}
  \centering
  \includegraphics[width=\linewidth]{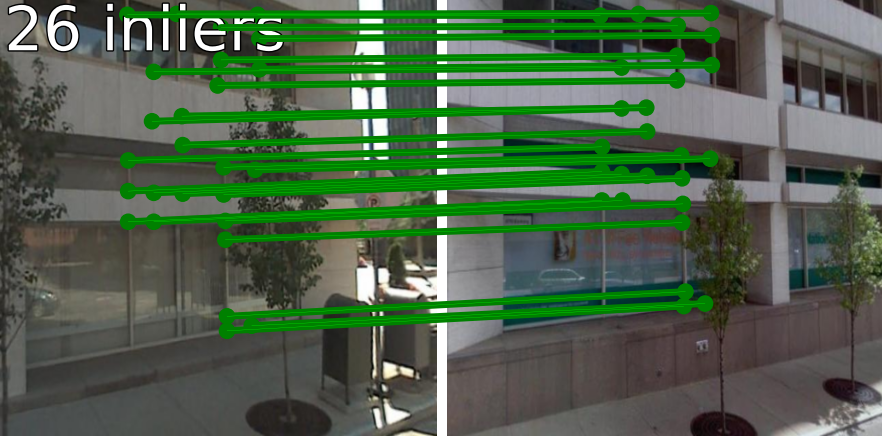}
  \caption{A query and its negative}
\end{subfigure}
\caption{\textbf{Example of a case when re-ranking through image matching fails.} The top-1 retrieved is shown next to the query on the left, and it's a positive. On the right, the top-2 retrieved image, which is a negative. SuperGlue + RANSAC finds fewer points in common between the pair on the left (only 7 inliers), and more between the wrong pair (26 inliers).}
\label{fig:hurtful_reranking}
\vspace{-3mm}
\end{figure}

\noindent Our contributions are as follows:
\begin{itemize}[noitemsep,topsep=1pt]
\item We conduct an extensive evaluation of state-of-the-art image matching methods for re-ranking in VPR, obtaining the most comprehensive benchmark up-to-date both in terms of methods and datasets;
\item Drawing from our comprehensive experimental results, we demonstrate the inadequacy of existing benchmarks in keeping up with the pace of research, showing that most of them are largely saturated, and provide insights on remaining challenges for future works;
\item We show that, contrary to common belief, in many cases re-ranking can worsen retrieval performance (see \cref{fig:hurtful_reranking}), and propose an approach to quantify prediction uncertainty in VPR using image matching methods, demonstrating that inlier counts provide a reliable measure of confidence for retrieval predictions.
\end{itemize}

\noindent By providing a perspective shift on modern retrieval pipelines, our work advances the state of the art in VPR and provides a foundation for future research in leveraging image matching methods for robust and reliable place recognition.

\section{Related Work}
\label{sec:formatting}

\myparagraph{Visual Place Recognition} (VPR)
is typically addressed as an image retrieval problem, leveraging a database of geo-tagged images \cite{Schubert2024-tutorial,Masone_2021_survey,Berton_2022_benchmark}. 
After the pioneering work of NetVLAD~\cite{Arandjelovic_2018_netvlad}, learned representations derived from deep networks became the de-facto standard; initially derived from CNNs \cite{Babenko_2014_neural_codes,Arandjelovic_2018_netvlad,Trivigno2023-divide,Schubert2024-tutorial} and, subsequently, transformer-based architectures \cite{Zhu_2023_r2former,Lu2023-sela,Lu2024-crica}. A key challenge addressed by these methods is the generation of compact, yet highly discriminative, global feature descriptors. Techniques for achieving this include various pooling strategies~\cite{Razavian_2015_mac,Tolias_2016_rmac,Radenovic_2019_gem, Arandjelovic_2018_netvlad, Mereu_2022_seqvlad, Hausler_2019, Nubert2021-HDC}, clustering-based feature aggregation~\cite{Zhang_2021_gated_netvlad,Kim_2017_crn,Peng_2021_appsvr,Izquierdo2024-salad}, MLP-based aggregations \cite{Alibey_2023_mixvpr}, and the adoption of a set of learnable tokens \cite{Alibey2024-BoQ}.
With the growing availability of geo-tagged images, modern research on VPR has moved towards efficient training protocols with stricter supervision, by leveraging curated datasets~\cite{Alibey_2022_gsvcities}, co-visibility constraints~\cite{Leyvavallina_2021_gcl} and class-based partitions of the database~\cite{Berton_2022_cosPlace,Berton2023-eigenplaces}.
A recent breakthrough in VPR \cite{Lu2023-sela,Keetha2024-anyloc} has been the adoption of vision foundation models such as DINOv2~\cite{oquab2023-dinov2}. Combining these training techniques and foundation models with optimal transport aggregation~\cite{Izquierdo2024-salad} and novel mining techniques~\cite{Izquierdo2024-clique, Berton_2025_megaloc} has led to methods with exceptional generalization capabilities.
We show that these recent advances allow to consider many long-standing VPR benchmarks as solved.

\myparagraph{Keypoint Detection and Description}
Finding repeatable keypoints (and associated descriptors) in an image is a longstanding problem of computer vision. 
Early handcrafted approaches adopted a detect-then-describe approach, typically based on local derivatives of the image~\cite{Lowe_2004_sift, Bay_2008_surf, artal_2015_orb}. 
With the advent of deep learning, learning-based approaches gained popularity. Pioneering works \cite{serra_2015_desc, tian_2017_l2net} employed contrastive learning to learn local descriptors with Convolutional Networks. 
SuperPoint~\cite{DeTone_2018_superpoint} proposed to generate synthetic shapes to train a neural network via self-supervision.
Subsequent works introduced a joint detect-and-describe paradigm, in which keypoints are implicitly defined as local maxima of the extracted features \cite{Dusmanu_2019_D2Net, Revaud_2019_r2d2, Gleize_2023_ICCV, Tyszkiewicz_2020_disk, Zhao_2023_aliked}.
More recently, DeDoDe~\cite{Edstedt_2023_dedode} proposes to separately optimize detection and description, in order to improve repeatability by enforcing 3D consistency constraints. Its follow-up Steerers~\cite{Bokman_2024_steerers} introduces rotation invariant descriptors, enabling several space and medical applications \cite{Berton_2024_EarthMatch, pielawski_2020_comir, Stoken_2023_CVPR}.

\myparagraph{Image Matching} aims at establishing pixel correspondences between different views of a scene. 
Traditionally, matches were established by finding mutual nearest neighbor on local keypoint descriptors \cite{Revaud_2019_r2d2}. This strategy can lead to errors as it does not allow reasoning on the global image context. A possible solution is to geometrically verify matches with RANSAC \cite{fischler_1981_random}, or to employ a learnable matcher such as SuperGlue \cite{Sarlin_2020_superglue}, 
a graph neural network-based approach. While SuperGlue operates a-posteriori of the matching stage,  LoFTR\cite{Sun_2021_loftr} foregoes the detection stage and proposes a detector-free paradigm where image context is incorporated thanks to the global attention mechanism of transformer architectures, thus improving robustness to repetitive patterns and low-texture areas. Following LoFTR, other methods adopted a detector-free paradigm \cite{Wang_2022_matchformer, Tang_2022_quadtree, huang_2023_adaptive, chen_2022_aspanformer, Bokman_2022_se2loftr, Zhou_2021_patch2pix}.
Alternatively, methods for dense feature matching aim to estimate every matchable pixel pair to obtain a dense warping of the two images \cite{Edstedt_2023_dkm, Edstedt_2024_roma}.
All these methods cast the matching problem in 2D, \ie without explicitly accounting for the geometrical properties of the scene. Recently, Dust3r \cite{Wang_2024_dust3r} and its follow-up Mast3r \cite{Leroy_2024_grounding}, propose to ground matches in 3D, by solving the task of 3D reconstruction from uncalibrated images, and then recovering point correspondences.
We conduct a comprehensive benchmark spanning a wide variety of image matching methods applied to re-ranking in VPR, identifying those most suited for the task. We introduce a methodology to automatically assess their potential to enhance retrieval accuracy, and propose a framework to quantify the uncertainty inherent in retrieval through image matching.

\myparagraph{Uncertainty Estimation} 
In VPR, naive uncertainty estimation could be obtained directly from the image retrieval model, through the L2-distance in feature space between the query and its nearest neighbors. To improve upon this simple baseline, several techniques have been proposed to explicitly model uncertainty. 
Examples include STUN~\cite{Cai_2022_stun} and BTL~\cite{Warburg_2021_bayesian}, which predict aleatoric uncertainty based only on the query’s image content, and SUE~\cite{Zaffar_2024_estimation}, which leverages the geographical distribution of the retrieved shortlist of candidates.
\section{Datasets}
To provide a comprehensive evaluation of the performance of image matching methods for uncertainty estimation and re-ranking in Visual Place Recognition, we use 10 datasets that span a broad spectrum of real-world scenarios, including outdoor and indoor environments, viewpoint variations, seasonal or weather changes, occlusions and day-to-night appearance shifts.

For \textbf{indoor environments}, we use the Baidu~\cite{Sun_2017_dataset} test set, which contains images captured in a mall with varying camera poses. This dataset features challenges such as perceptually aliased structures and distractors (e.g., people), making it ideal for VPR evaluation.

In the domain of \textbf{medium-scale} (10k-100k database size) \textbf{urban VPR}, we employ three widely used datasets: the validation set of MSLS~\cite{Warburg_2020_msls}, and the test sets of Pitts30k~\cite{Arandjelovic_2018_netvlad} and Tokyo 24/7~\cite{Torii_2018_tokyo247}. MSLS consists of over 1 million images from multiple cities and the ones from San Francisco and Copenaghen are used as validation set. Pitts30k, built from Google StreetView images of Pittsburgh, includes 6,816 test queries from different years and is often used as a benchmark in VPR literature. 
Tokyo 24/7 presents a set of 315 queries from smartphone photos taken in central Tokyo at day, sunset, and night. Thus, it is suitable to assess performance under \textit{varying lighting conditions}.

For \textbf{large-scale urban VPR}, we use the San Francisco eXtra Large (SF-XL)~\cite{Berton_2022_cosPlace} dataset, which contains over 41 million images. The SF-XL test set includes 2.8 million images with multiple query sets. The official test sets, V1 and V2, assess \textit{viewpoint changes} and \textit{domain shifts}, with images sourced from Flickr and smartphones, respectively. Additionally, the SF-XL Night and SF-XL Occlusion~\cite{Barbarani_2023_local} queries introduce further challenges, with \textit{night-time} imagery and images featuring heavy \textit{occlusions} like cars and pedestrians.

Lastly, to evaluate VPR in \textbf{diverse weather conditions}, we use the SVOX~\cite{Berton_2021_svox} dataset, which provides a robust test set for cross-domain VPR. The dataset spans Oxford, UK, using Google StreetView for the database and the Oxford RobotCar~\cite{Maddern_2017_robotCar} dataset for queries. Here, we select the queries from the Sun and Night subsets.

Table~\ref{tab:datasets} provides a summary of all the datasets, showing the number of queries and images in the database, along with the types of scenery and domain shifts.

\begin{table}
\begin{adjustbox}{width=\linewidth}
\centering
\begin{tabular}{lcccc}
\toprule
Dataset & \# Queries & \# Database Images & Scenery & Domain Shift  \\
\hline
Baidu & 3k & 5k & Indoor & Viewpoint Shift/Occlusions \\
MSLS Val & 11k & 19k & Urban & Day-Night \\
Pitts30k & 7k & 10k & Urban & None \\
SF-XL Night & 466 & 2.8M & Urban & Day-Night \\
SF-XL Occlusion & 76 & 2.8M & Urban & Occlusions \\
SF-XL test V1 & 1000 & 2.8M & Urban & Viewpoint / Night \\
SF-XL test V2 & 598 & 2.8M & Urban & Viewpoint \\
SVOX Night & 823 & 17k & Urban & Day-Night \\
SVOX Sun & 854 & 17k & Urban & Weather \\
Tokyo 24/7 & 315 & 76k & Urban & Day-Night \\
\bottomrule
\end{tabular}
\end{adjustbox}
\vspace{-3mm}
\caption{\textbf{Datasets.} For each dataset, the number of queries and database images, scenery and types of domain shift in the test set is provided, except for MSLS, where the validation set is used instead.}
\vspace{-3mm}
\label{tab:datasets}
\end{table}

\section{Experiments}
\label{sec:experiments}

In this work we aim to understand if and when, given the current state of Visual Place Recognition (retrieval) models, image matching methods for re-ranking are still relevant and useful.
In the next section we compute and showcase results on such task.

\begin{figure}
\centering
\includegraphics[width=\linewidth]{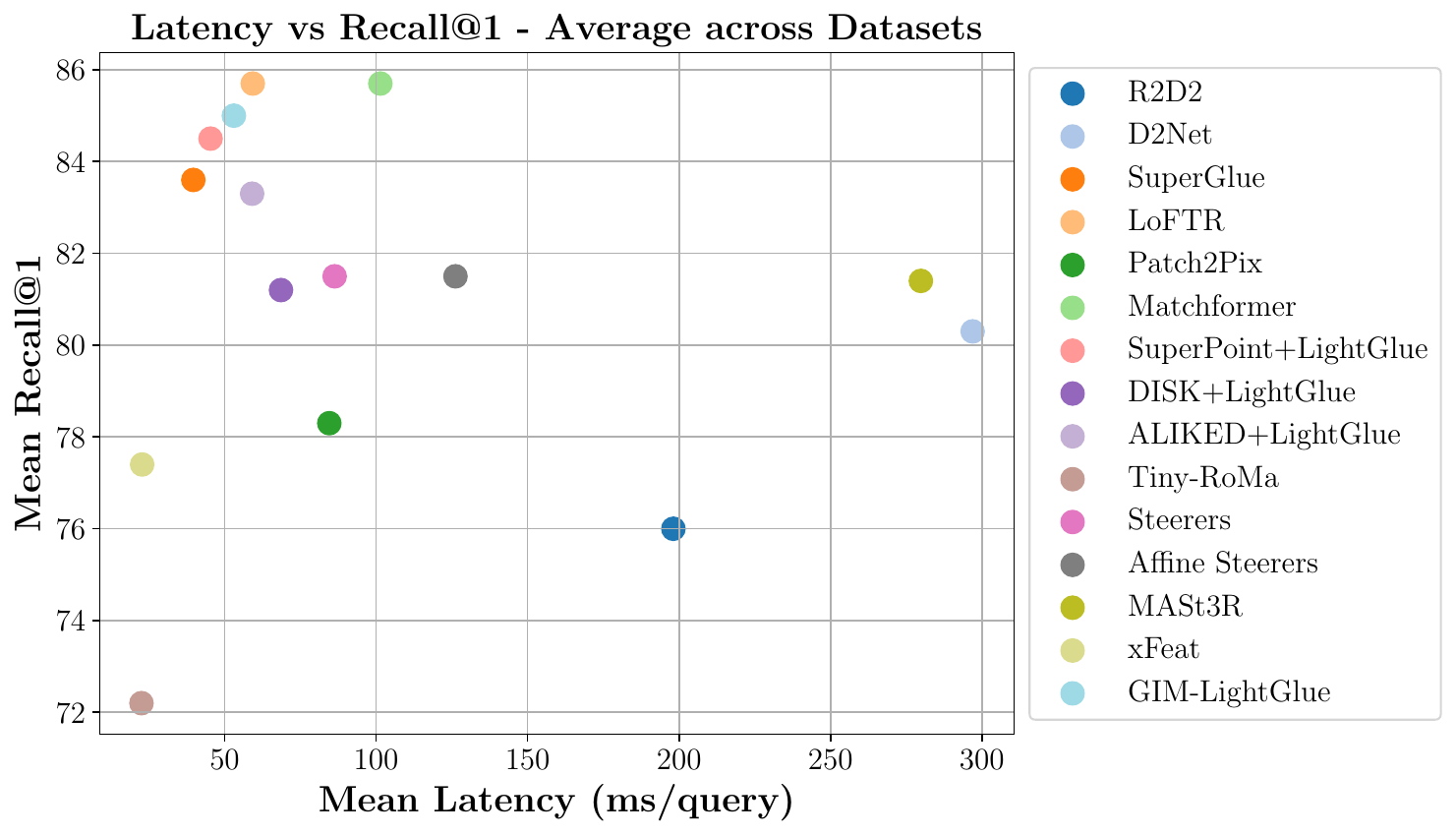}
\vspace{-7mm}
\caption{\textbf{Plot displaying the mean Recall@1 after re-ranking and mean latency for different methods.} The mean Recall@1 is computed over the datasets, while the mean latency is the average time to process each query over all datasets. The shortlist of candidates for the Recall@1 is obtained with MegaLoc and distance threshold fixed at 25 meters.}
\label{fig:mean_recall_latency}
\vspace{-3mm}
\end{figure}

\begin{table*}
\begin{adjustbox}{width=\linewidth}
\centering
\setlength{\tabcolsep}{2pt}
\begin{tabular}{lcccccccccccccccccccccccccccccc@{}|@{}ccc}
\toprule
\multirow{3}{*}{{\begin{tabular}[c]{@{}c@{}}Method\end{tabular}}} &
\multicolumn{2}{c}{Baidu} & & 
\multicolumn{2}{c}{MSLS Val} & & 
\multicolumn{2}{c}{Pitts30k} & & 
\multicolumn{2}{c}{SF-XL Night} & & 
\multicolumn{2}{c}{SF-XL Occlusion} & &
\multicolumn{2}{c}{SF-XL test V1} & &
\multicolumn{2}{c}{SF-XL test V2} & &
\multicolumn{2}{c}{SVOX Night} & &
\multicolumn{2}{c}{SVOX Sun} & &
\multicolumn{2}{c}{Tokyo 24/7} & & &
\multicolumn{2}{c}{Average} \\
& 
\multicolumn{2}{c}{R@100 = 99.9} & & 
\multicolumn{2}{c}{R@100 = 97.6} & & 
\multicolumn{2}{c}{R@100 = 99.6} & & 
\multicolumn{2}{c}{R@100 = 85.0} & & 
\multicolumn{2}{c}{R@100 = 92.1} & & 
\multicolumn{2}{c}{R@100 = 99.0} & & 
\multicolumn{2}{c}{R@100 = 99.0} & & 
\multicolumn{2}{c}{R@100 = 99.6} & & 
\multicolumn{2}{c}{R@100 = 99.9} & & 
\multicolumn{2}{c}{R@100 = 99.7} & & &
\multicolumn{2}{c}{R@100 = 97.1} \\
\cline{2-3} \cline{5-6} \cline{8-9} \cline{11-12} \cline{14-15} \cline{17-18} \cline{20-21} \cline{23-24} \cline{26-27} \cline{29-30} \cline{33-34}
& R@1 & R@10 & & R@1 & R@10 & & R@1 & R@10 & & R@1 & R@10 & & R@1 & R@10 & & R@1 & R@10 & & R@1 & R@10 & & R@1  & R@10 & & R@1  & R@10 & & R@1 & R@10 & & & R@1 & R@10 \\
\hline

- & 87.7 & 98.0 && \textbf{91.0} & \textbf{95.8} && \textbf{94.1} & \textbf{98.2} && 52.8 & 73.8 && 51.3 & 75.0 && \textbf{95.3} & 98.0 && \underline{94.8} & \underline{98.5} && \textbf{95.1} & \underline{98.8} && \textbf{96.5} & \underline{99.6} && 96.5 & \underline{99.4} &&& \underline{85.5} & 93.5 \\

\hline
R2D2 (NeurIPS '19) & 90.4 & 98.5 && 75.8 & 88.3 && 83.8 & 96.6 && 34.8 & 59.2 && 46.1 & 67.1 && 86.6 & 94.1 && 92.0 & 97.8 && 72.7 & 84.3 && 87.5 & 93.2 && 89.8 & 96.8 &&& 76.0 & 87.6 \\
D2Net (CVPR '19) & 91.4 & 98.8 && 78.1 & 91.2 && 86.2 & 97.2 && 50.6 & 71.5 && 52.6 & 75.0 && 90.7 & 96.8 && 93.1 & 98.2 && 78.4 & 94.0 && 88.5 & 96.7 && 93.7 & 97.8 &&& 80.3 & 91.7 \\
SuperGlue (CVPR '20) & 91.9 & 99.1 && 85.4 & 94.3 && 85.5 & 97.2 && \underline{60.7} & 75.5 && 55.3 & \underline{81.6} && 93.4 & \underline{98.2} && 90.8 & 98.3 && 85.7 & 97.1 && 91.3 & 98.1 && 95.9 & \underline{99.4} &&& 83.6 & \underline{93.9} \\
LoFTR (CVPR '21) & \textbf{94.8} & 98.8 && 85.4 & 93.9 && 87.2 & 96.3 && 58.2 & \underline{75.8} && \textbf{60.5} & 75.0 && 93.7 & 97.1 && 94.1 & 98.3 && 91.0 & 98.2 && \underline{95.9} & 99.2 && 96.5 & \textbf{99.7} &&& \textbf{85.7} & 93.2 \\
Patch2Pix (CVPR '21) & 91.1 & 99.0 && 79.7 & 91.1 && 84.9 & 96.4 && 38.0 & 61.2 && 52.6 & 68.4 && 89.9 & 96.3 && 92.5 & 98.0 && 67.3 & 89.1 && 92.3 & 98.0 && 94.9 & 98.7 &&& 78.3 & 89.6 \\
Matchformer (ACCV '22) & 93.7 & 98.8 && 83.6 & 91.9 && 88.2 & 97.1 && \textbf{61.8} & \underline{75.8} && 55.3 & 71.1 && 94.0 & 97.4 && \textbf{95.5} & 98.3 && \underline{92.6} & 98.4 && 95.7 & 99.2 && 96.8 & 99.0 &&& \textbf{85.7} & 92.7 \\
SuperPoint+LightGlue (ICCV '23) & 92.6 & \textbf{99.3} && 83.9 & 93.5 && 85.9 & 97.3 && 58.8 & \textbf{76.4} && \underline{56.6} & \textbf{84.2} && 94.3 & \underline{98.2} && 91.6 & 98.2 && 89.4 & 96.8 && 93.8 & 99.3 && \textbf{98.4} & 99.0 &&& 84.5 & \textbf{94.2} \\
DISK+LightGlue (ICCV '23) & 91.4 & 99.1 && 84.6 & 93.6 && 84.6 & 96.7 && 50.2 & 71.9 && 53.9 & \underline{81.6} && 90.7 & 97.7 && 91.5 & \textbf{98.7} && 80.8 & 89.8 && 90.7 & 97.1 && 94.0 & \textbf{99.7} &&& 81.2 & 92.6 \\
ALIKED+LightGlue (ICCV '23) & \underline{94.3} & \underline{99.2} && \underline{87.5} & 94.4 && 85.2 & 97.4 && 56.2 & 73.8 && \underline{56.6} & 77.6 && 92.8 & \textbf{98.4} && 91.5 & 98.2 && 83.0 & 92.0 && 88.6 & 96.1 && \underline{97.5} & \underline{99.4} &&& 83.3 & 92.7 \\
RoMa (CVPR '24) & 88.7 & 98.3 && 45.9 & 88.2 && 72.8 & 95.9 && 44.6 & 74.9 && 36.8 & 77.6 && 88.0 & 96.9 && 88.5 & 98.0 && 65.1 & 98.3 && 71.4 & 93.3 && 84.1 & \textbf{99.7} &&& 68.6 & 92.1 \\
Tiny-RoMa (CVPR '24) & 88.7 & 98.3 && 76.2 & 91.6 && 81.8 & 96.7 && 41.2 & 69.1 && 50.0 & 71.1 && 88.0 & 96.9 && 88.5 & 98.0 && 48.8 & 85.7 && 71.4 & 93.3 && 87.3 & 98.4 &&& 72.2 & 89.9 \\
Steerers (CVPR '24) & 93.1 & 98.7 && 77.0 & 87.7 && 85.1 & 96.8 && 48.5 & 67.2 && 53.9 & 76.3 && 92.0 & 97.6 && 91.5 & 98.2 && 82.7 & 95.3 && 93.3 & 98.1 && \underline{97.5} & \textbf{99.7} &&& 81.5 & 91.6 \\
Affine Steerers (ECCV '24) & 91.3 & 98.0 && 79.8 & 90.9 && 85.1 & 96.8 && 50.4 & 70.2 && 53.9 & 76.3 && 91.3 & 97.5 && 91.5 & 98.2 && 82.7 & 95.3 && 92.5 & 97.8 && 96.2 & 98.4 &&& 81.5 & 91.9 \\
DUSt3R (CVPR '24) & 85.0 & 98.2 && 63.0 & 80.4 && 79.5 & 93.5 && 35.6 & 56.2 && 42.1 & 63.2 && 78.6 & 92.7 && 66.1 & 93.3 && 60.3 & 69.9 && 71.7 & 82.2 && 86.0 & 93.7 &&& 66.8 & 82.3 \\
MASt3R (ECCV '24) & 89.8 & 99.1 && 71.7 & 93.0 && 85.9 & \underline{98.0} && 56.2 & 74.2 && \underline{56.6} & 80.3 && 90.4 & \underline{98.2} && 83.4 & 98.0 && \underline{92.6} & \textbf{99.1} && 93.4 & \textbf{99.8} && 94.0 & \textbf{99.7} &&&  81.4 & \underline{93.9} \\
xFeat (CVPR '24) & 86.8 & 97.8 && 83.0 & 92.1 && 86.6 & 96.9 && 45.3 & 67.6 && 44.7 & 75.0 && 88.7 & 96.5 && 91.1 & 98.3 && 75.2 & 90.6 && 83.5 & 95.2 && 88.9 & 98.4 &&& 77.4 & 90.8 \\
GIM-DKMv3 (ICLR '24) & 41.6 & 94.8 && 4.4 & 35.0 && 40.1 & 91.4 && 31.1 & 71.7 && 26.3 & 73.7 && 33.8 & 88.2 && 46.0 & 94.8 && 29.6 & 89.6 && 21.8 & 86.1 && 47.9 & 97.5 &&& 32.3 & 82.3 \\
GIM-LightGlue (ICLR '24) & 92.4 & 98.8 && 86.4 & \underline{94.5} && \underline{88.8} & 97.6 && 59.7 & 74.0 && 53.9 & 77.6 && \underline{94.4} & \underline{98.2} && 92.1 & 98.3 && 91.0 & 96.5 && 94.7 & 98.4 && 96.5 & 99.0 &&& 85.0 & 93.3 \\
\bottomrule
\end{tabular}
\end{adjustbox}
\vspace{-3mm}
\caption{\textbf{Recalls before and after applying re-ranking.} Recalls are computed by setting the distance threshold to 25 meters.
The shortlist of candidates to be re-ranked is obtained with MegaLoc, and the results with such shortlist are shown in the first row.
Re-ranking has been applied to the first 100 candidates (\ie $K=100$). Next to each dataset's name, we show the R@100, which in practice sets the upper bound of the maximum recalls achievable after re-ranking. Best results are in \textbf{bold}, second best are \underline{underlined}.
}
\vspace{-3mm}
\label{tab:reranking_results_megaloc_25m}
\end{table*}

\subsection{Re-ranking}
\label{sec:exp_reranking}

\myparagraph{Implementation Details:} The count of inliers (\ie matches that ``survive'' the post-processing with RANSAC) can be leveraged to re-rank the candidate shortlist obtained through retrieval methods, thereby enhancing the Recall@$K$ metric.

We use the state-of-the-art MegaLoc~\cite{Berton_2025_megaloc} as the retrieval model across all datasets. Unless explicitly stated, we follow the standard practices in Visual Place Recognition, considering an image $I$ retrieved from the database a correct match for the query $q$ if and only if their locations are at most 25 meters apart. Formally, the prediction provided by $I$ is correct if $d_{g}(q, I) \le 25$, where $d_{g}$ denotes the geographic distance expressed in meters. Each input image is resized to $322 \times 322$ pixels before being processed by MegaLoc. We compare the various image matching methods by using their default hyper-parameters and resizing each image to $512 \times 512$ pixels.

To evaluate the re-ranking performance of the image matching methods, the top 100 nearest neighbors for each query are initially retrieved from the database using MegaLoc. The re-ranking process then sorts these 100 candidate images based on the number of inliers $i_{q}^{(j)}$ between the query $q$ and the $j$-th nearest neighbor, for $j = 1, 2, \ldots, 100$, in descending order.

\myparagraph{Image matching methods:} 
For this analysis, we selected a substantial number of open-source image matching models\footnote{methods available in the Image Matching Models GitHub repository \cite{Berton_2024_EarthMatch} at \url{https://github.com/alexstoken/image-matching-models}}: R2D2~\cite{Revaud_2019_r2d2}, D2Net~\cite{Dusmanu_2019_D2Net}, SuperGlue~\cite{Sarlin_2020_superglue}, LoFTR~\cite{Sun_2021_loftr}, Patch2Pix~\cite{Zhou_2021_patch2pix}, Matchformer~\cite{Wang_2022_matchformer}, SuperPoint+LightGlue~\cite{DeTone_2018_superpoint,Lindenberger_2023_lightglue}, DISK+LightGlue~\cite{Tyszkiewicz_2020_disk,Lindenberger_2023_lightglue}, ALIKED+LightGlue~\cite{Zhao_2023_aliked,Lindenberger_2023_lightglue}, RoMa and Tiny-RoMa~\cite{Edstedt_2024_roma}, Steerers~\cite{Bokman_2024_steerers}, Affine Steerers~\cite{Bokman_2024_affine}, DUSt3R~\cite{Wang_2024_dust3r}, MASt3R~\cite{Leroy_2024_grounding}, xFeat~\cite{Potje_2024_xfeat}, GIM-DKMv3~\cite{Edstedt_2023_dkm,Shen_2024_gim} and GIM-LightGlue~\cite{Shen_2024_gim,Lindenberger_2023_lightglue}.

\myparagraph{Baseline:}  The baseline is represented by the pure retrieval performance of MegaLoc \cite{Berton_2025_megaloc}. MegaLoc is trained on a dataset made up of train sets from SF-XL \cite{Berton_2022_cosPlace}, GSV-Cities \cite{Alibey_2022_gsvcities}, MSLS \cite{Warburg_2020_msls}, MegaScenes \cite{Tung_2024_Megascenes} and ScanNet \cite{Dai_2017_scannet}.

\myparagraph{Evaluation Metric:}
The evaluation is conducted using Recall@$K$ at a fixed distance threshold $\tau$. Recall@$K$ measures the percentage of queries for which at least one of the top-$K$ retrieved images is within $\tau$ meters of the query's ground-truth location. Unless otherwise specified, experiments are carried out with $\tau = 25$. A higher value of Recall@$K$ relative to MegaLoc’s performance indicates better re-ranking capability of the image matching method.

\myparagraph{Results:}
Table~\ref{tab:reranking_results_megaloc_25m} presents the Recall@1 and Recall@10 values after the re-ranking process, along with MegaLoc's performance for each dataset. Since re-ranking is applied to the top 100 retrieved images, the Recall@100 (shown in the table's header) represents the upper bound on performance that can be achieved through re-ranking.

An intriguing observation from our results is that, contrary to previous findings in the literature \cite{Wang_2022_TransVPR, hausler2024pairvpr, Barbarani_2023_local, Hausler_2021_patch_netvlad, Lu2023-sela, Tzachor_2024_effoVPR}, applying re-ranking does not universally enhance performance: we believe this to be due to recent improvements in the VPR literature (\eg our retrieval baseline MegaLoc), which provide good results that are hard to improve upon by means of re-ranking.
Specifically, it is the case of Pitts30k, MSLS, SVOX and SF-XL, where even the best re-ranking methods cause a drop in R@1.
At the same time, MegaLoc essentially saturates these long-standing benchmarks through pure retrieval alone.
This finding challenges the widely held belief that re-ranking consistently refines initial matches, and motivates us to further investigate when image matching methods can prove beneficial, rather than assuming their unequivocal benefit.

For the datasets that present several occlusions in the query sets, namely Baidu and SF-XL Occlusion, image matching methods--on average--are able to improve the Recall@1 by 3.3\% and 0.5\%, respectively.
However, when considering the average performance across all datasets, only two methods—LoFTR and Matchformer—improve Recall@1, while three methods—SuperPoint+LightGlue, SuperGlue, and MASt3R—improve Recall@10. No single method enhances both metrics. This suggests that image matching methods are particularly beneficial when the retrieval model performs poorly and struggles to accurately map retrieved images in its output space. A prime example is SF-XL Night, where Matchformer improves Recall@1 by 9\% and Recall@10 by 3\%, and seven methods, in total, assist in re-ranking the candidate shortlist for both recall values.

Figure~\ref{fig:mean_recall_latency} offers a comparison of the analyzed methods, showing the average Recall@1 alongside the time taken to process a single query (\ie re-ranking its top-100 predictions). This includes extracting keypoints for both the query and database image, and matching inliers. Ideally, methods that are both accurate and time-efficient are best suited for real-time re-ranking applications.

\begin{table*}
\begin{center}
\begin{adjustbox}{width=0.75\linewidth}
\centering
\setlength{\tabcolsep}{2pt}
\begin{tabular}{lcccccccccc|c}
\toprule
\multirow{2}{*}{Method} & \multirow{2}{*}{Baidu} & MSLS & \multirow{2}{*}{Pitts30k} & SF-XL & SF-XL & SF-XL & SF-XL & SVOX & SVOX & Tokyo & \multirow{2}{*}{Average} \\
& & Val & & Night & Occlusion & test V1 & test V2 & Night & Sun & 24/7 & \\
\hline

L2-distance & 94.0 & 97.0 & \textbf{99.1} & 69.8 & \underline{77.5} & \underline{99.5} & 98.0 & 99.2 & \underline{99.1} & \textbf{99.9} & 93.3 \\
PA-Score & 93.8 & 96.5 & 98.9 & 67.3 & 71.6 & 98.6 & 98.0 & 99.0 & 98.9 & 99.8 & 92.2 \\
SUE & \underline{95.5} & \underline{97.1} & 98.6 & \underline{73.6} & 73.5 & 99.1 & \textbf{98.2} & \underline{99.6} & 99.0 & \textbf{99.9} & \underline{93.4} \\
Random & 88.0 & 90.8 & 94.3 & 53.2 & 45.9 & 94.7 & 96.0 & 94.8 & 97.6 & 96.9 & 85.2 \\
\hline
R2D2 (NeurIPS '19) & 96.5 & 96.4 & 98.4 & 69.7 & 80.5 & 99.5 & 97.8 & 99.1 & 99.6 & 99.7 & 93.7 \\
D2Net (CVPR '19) & 97.3 & 96.2 & 98.4 & 73.2 & 78.2 & \textbf{99.7} & 97.7 & 99.2 & 99.4 & 99.8 & 93.9 \\
SuperGlue (CVPR '20) & \textbf{97.4} & 97.2 & 98.8 & 75.5 & 84.2 & \textbf{99.7} & 97.7 & 99.5 & 99.4 & \textbf{99.9} & \textbf{94.9} \\
LoFTR (CVPR '21) & 97.3 & 97.0 & 98.8 & 73.9 &\textbf{84.5} & \textbf{99.7} & 98.0 & 99.5 & 99.7 & \textbf{99.9} & 94.8 \\
Patch2Pix (CVPR '21) & 96.6 & 96.7 & 98.6 & 73.5 & 78.0 & \textbf{99.7} & 97.9 & 99.1 & 99.4 & \textbf{99.9} & 93.9 \\
Matchformer (ACCV '22) & 97.2 & 96.9 & 98.9 & 74.9 & 83.1 & \textbf{99.7} & \underline{98.1} & 99.6 & 99.6 & \textbf{99.9} & 94.8 \\
SuperPoint+LightGlue (ICCV '23) & 97.3 & 97.4 & 98.9 & 75.8 & 82.4 & 99.6 & 97.7 & 99.5 & 99.5 & \textbf{99.9} & 94.8 \\
DISK+LightGlue (ICCV '23) & 95.7 & 97.0 & 98.9 & 70.8 & 80.9 & 99.3 & 96.9 & 99.2 & 99.3 & 99.8 & 93.8 \\
ALIKED+LightGlue (ICCV '23) & 96.9 & \textbf{97.6} & 99.0 & 70.6 & 79.6 & 99.6 & 97.8 & 99.5 & 99.5 & \textbf{99.9} & 94.0 \\
RoMa (CVPR '24) & 94.8 & 96.1 & 96.7 & 62.1 & 72.5 & 99.3 & 94.8 & 99.4 & 99.7 & 99.8 & 91.5 \\
Tiny-RoMa (CVPR '24) & 97.1 & 96.5 & 99.0 & 69.4 & 80.7 & 99.6 & 97.0 & 98.6 & 99.4 & 99.8 & 93.7 \\
Steerers (CVPR '24) & 96.8 & 96.8 & 99.0 & 72.5 & 79.0 & 99.4 & 96.8 & 99.4 & 99.3 & 99.8 & 93.9 \\
Affine Steerers (ECCV '24) & 96.5 & 96.8 & 98.5 & 69.7 & 81.2 & 99.3 & 97.1 & 99.3 & 99.3 & \textbf{99.9} & 93.8 \\
DUSt3R (CVPR '24) & 95.3 & 97.0 & 98.7 & 63.8 & 59.1 & 98.0 & 94.6 & 98.6 & 98.7 & 99.2 & 90.3 \\
MASt3R (ECCV '24) & 95.7 & 96.8 & \textbf{99.1} & 67.4 & 79.2 & 99.6 & 96.1 & \textbf{99.8} & \textbf{99.8} & \textbf{99.9} & 93.3 \\
xFeat (CVPR '24) & 95.7 & 96.8 & 98.6 & 74.1 & 79.3 & 99.6 & 97.1 & 98.7 & 99.3 & 99.7 & 93.9 \\
GIM-DKMv3 (ICLR '24) & 92.6 & 92.4 & 94.5 & 62.2 & 63.4 & 96.6 & 93.8 & 97.9 & 98.7 & 99.6 & 89.2 \\
GIM-LightGlue (ICLR '24) & 97.1 & 97.3 & 98.9 & \textbf{76.5} & 80.4 & 99.6 & 98.0 & 99.6 & 99.5 & \textbf{99.9} & 94.7 \\
\bottomrule
\end{tabular}
\end{adjustbox}
\end{center}
\vspace{-3mm}
\caption{\textbf{The AUPRC of all the baselines and image matching methods}, split according to group type. The shorlist of candidates is obtained with MegaLoc. Distance threshold is fixed at 25 meters. Best
overall results on each dataset are in \textbf{bold}, best results for each group are \underline{underlined}.}
\label{tab:auprc_megaloc_25m}
\vspace{-3mm}
\end{table*}
\begin{figure*}
\centering
\begin{subfigure}{0.33\textwidth}
  \centering
  \includegraphics[width=\linewidth]{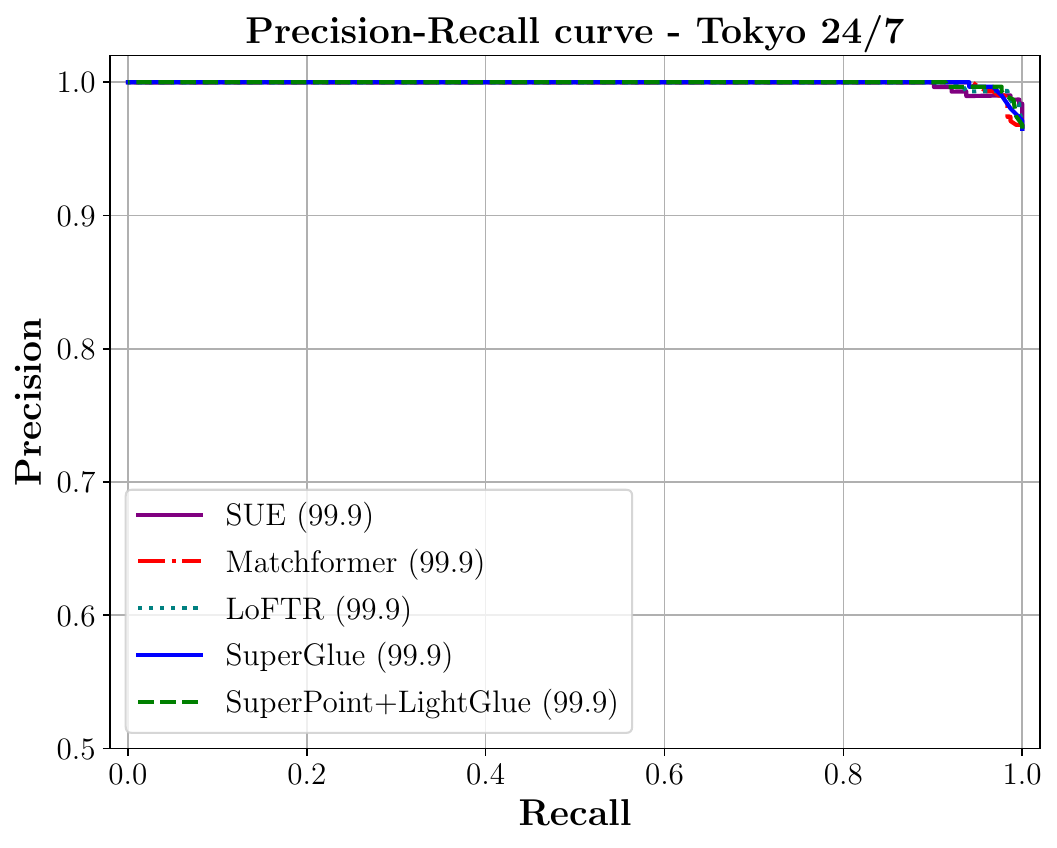}
  \label{fig:pr_curve_tokyo247}
\end{subfigure}%
\begin{subfigure}{0.33\textwidth}
  \centering
  \includegraphics[width=\linewidth]{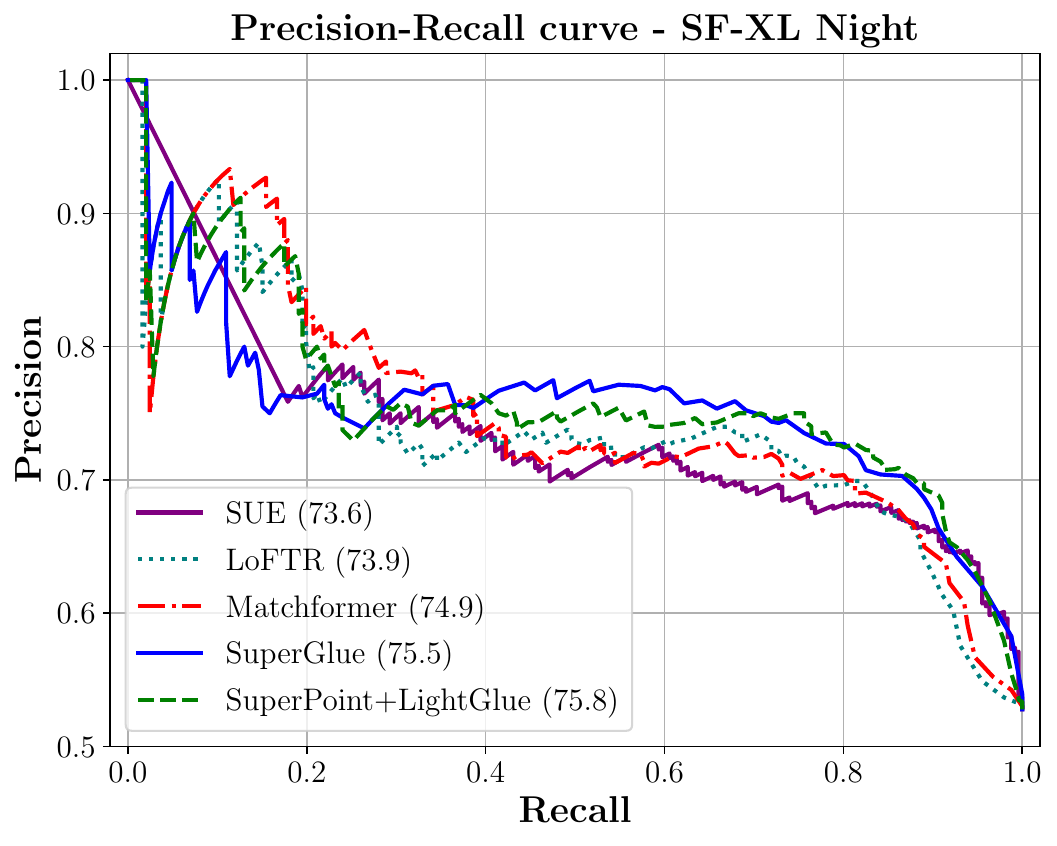}
  \label{fig:pr_curve_sf_xl_night}
\end{subfigure}
\begin{subfigure}{0.33\textwidth}
  \centering
  \includegraphics[width=\linewidth]{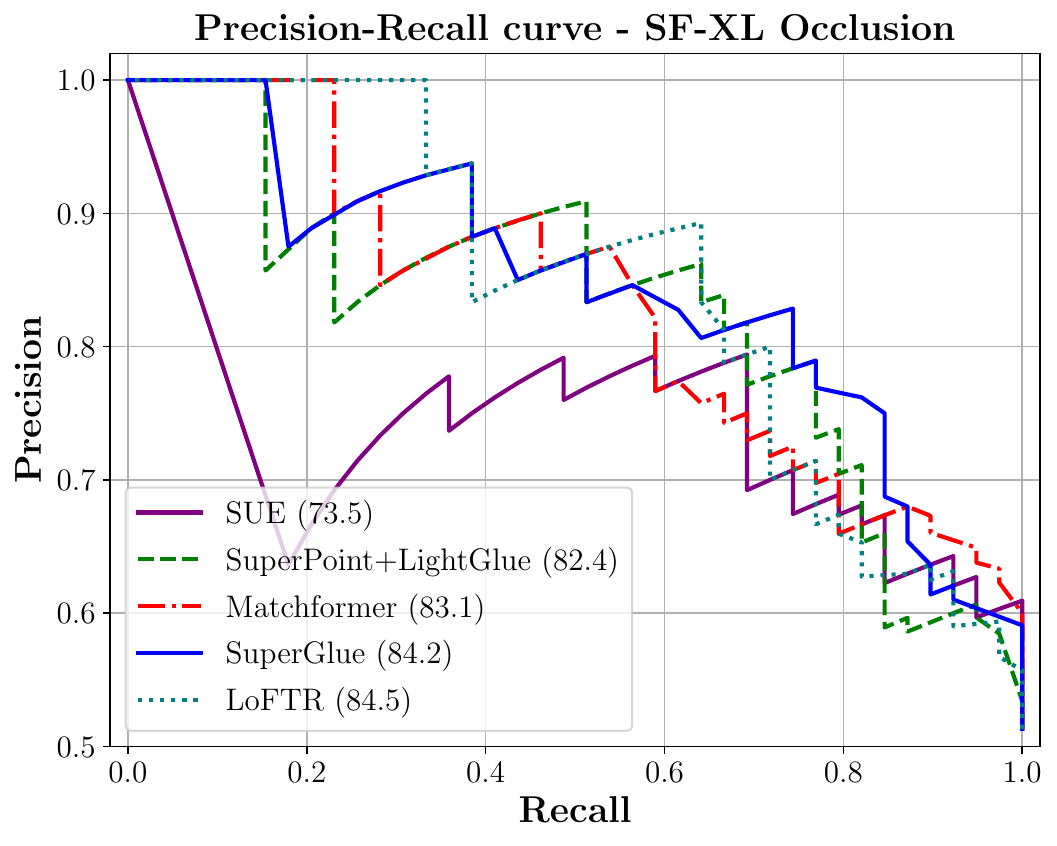}
  \label{fig:pr_curve_sf_xl_}
\end{subfigure}
\vspace{-5mm}
\caption{\textbf{Precision-Recall curves}, computed for the top-4 image matching methods on Tokyo 24/7, SF-XL Night, and SF-XL Occlusion, together with SUE, which is representative of the baselines when the shortlist of candidates is obtained with MegaLoc. Distance threshold is fixed at 25 meters.}
\label{fig:precision_recall_curves_megaloc}
\vspace{-3mm}
\end{figure*}
\begin{figure}
\centering
\includegraphics[width=0.9\linewidth]{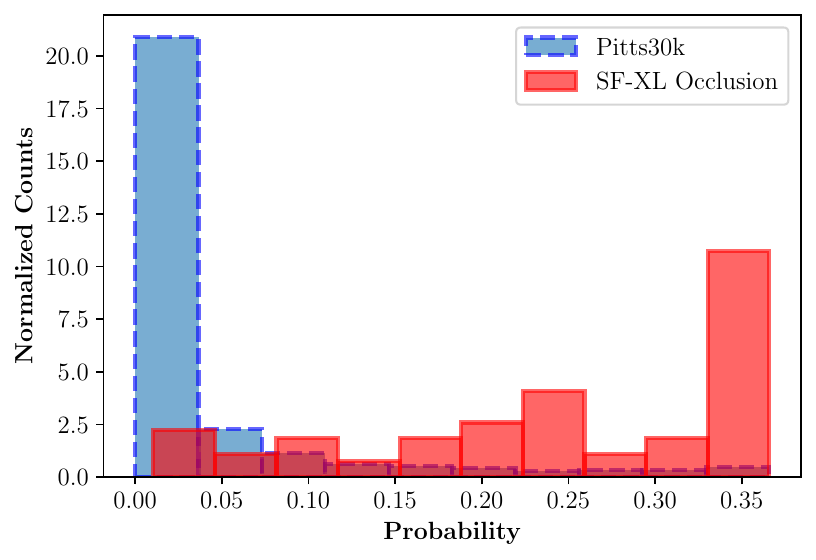}
\vspace{-3mm}
\caption{\textbf{Histogram of probabilities of being a wrongly localized query}, \ie with a top-1 prediction further than 25 meters.
The probabilities are computed on the query sets of Pitts30k and SF-XL Occlusion using a Logistic Regression trained on the uncertainty scores produced by MASt3R on MSLS Val.}
\label{fig:histogram_prob}
\vspace{-3mm}
\end{figure}

\subsection{Prediction Uncertainty via Image Matching}
\label{sec:exp_uncertainty}

Results from the previous section show that re-ranking can prove detrimental for performances in cases where the retrieval R@1 is near 100\%. 
However, in real-world scenarios, there is no such concept as a saturated dataset, as queries are fed to the system individually, and can potentially come from different data distributions.
Therefore, it is important to estimate which queries can be solved by retrieval alone, and which ones can benefit from re-ranking.
To this end, we posit that, given a reasonable estimation of uncertainty, we can find a correlation between uncertainty and potential improvements attainable via re-ranking.
In the rest of this subsection, we aim to validate our hypothesis. Namely, to understand whether a reliable uncertainty value exists (\ie the probability that a given query has been wrongly localized), and whether such value is effectively correlated with the impact that re-ranking has on a given query.
Simply put, we aim to verify that, in order to maximize performance, in a real-world application we could apply re-ranking only for high-uncertainty queries, whereas high-confidence predictions can be left untouched in order not to jeopardize positive results.

\myparagraph{Baselines:}
The topic of uncertainty estimation  has been studied for image retrieval, either by directly learning to predict aleatoric uncertainty at training time \cite{Warburg_2021_bayesian}, or through post-hoc techniques at inference \cite{Zaffar_2024_estimation}.
Among the latter, a simple technique entails using the L2-distance to the nearest neighbor for each query, $u_{q} \triangleq d_{(1)}$, and the perceptual aliasing score (PA-score), \ie the ratio of the distances between the first and second nearest neighbors in the database, \ie, $u_{q} \triangleq \frac{d_{(1)}}{d_{(2)}}$. Additionally, we include SUE~\cite{Zaffar_2024_estimation}, the state-of-the-art method for uncertainty estimation in the VPR task, which considers the geographic spread of the shortlist of candidates retrieved by MegaLoc.
We further introduce a Random baseline, where uncertainty scores are sampled from a uniform distribution, $u_{q} \sim \mathcal{U}(0, 1)$.

Besides these methods that focus purely on uncertainty estimation for VPR, image matching models have been shown to provide good results for the task \cite{Zaffar_2024_estimation}: intuitively, when a retrieved prediction has few matches with the given query, the uncertainty will be high, whereas in the presence of numerous matches between two images we can confidently state that the two represent the same place.

\myparagraph{Image matching for uncertainty estimation:}
To quantify the uncertainty associated with each method, we measure the number of inliers $i_{q}^{(1)}$ between the query $q$ and the nearest neighbor $I_{(1)}$ (with corresponding L2-distance in the output space of MegaLoc indicated as $d_{(1)}$). The uncertainty is then defined as $u_{q} \triangleq - i_{q}^{(1)}$, with fewer inliers indicating greater uncertainty.

\myparagraph{Evaluation Metrics:} We adopt the evaluation framework of previous uncertainty for VPR papers \cite{Zaffar_2024_estimation} across all datasets. The evaluation metric is the Area Under the Precision-Recall Curve (AUPRC), where a higher value indicates better discrimination between correct and incorrect queries based on uncertainty scores.

\begin{table*}
\begin{adjustbox}{width=\linewidth}
\centering
\setlength{\tabcolsep}{2pt}
\begin{tabular}{lcccccccccccccccccccccccccccccc@{}|@{}ccc}
\toprule
\multirow{3}{*}{{\begin{tabular}[c]{@{}c@{}}Method\end{tabular}}} &
\multicolumn{2}{c}{Baidu} & & 
\multicolumn{2}{c}{MSLS Val} & & 
\multicolumn{2}{c}{Pitts30k} & & 
\multicolumn{2}{c}{SF-XL Night} & & 
\multicolumn{2}{c}{SF-XL Occlusion} & &
\multicolumn{2}{c}{SF-XL test V1} & &
\multicolumn{2}{c}{SF-XL test V2} & &
\multicolumn{2}{c}{SVOX Night} & &
\multicolumn{2}{c}{SVOX Sun} & &
\multicolumn{2}{c}{Tokyo 24/7} & & &
\multicolumn{2}{c}{Average} \\
& 
\multicolumn{2}{c}{R@100 = 100.0} & & 
\multicolumn{2}{c}{R@100 = 98.4} & & 
\multicolumn{2}{c}{R@100 = 100.0} & & 
\multicolumn{2}{c}{R@100 = 92.3} & & 
\multicolumn{2}{c}{R@100 = 98.7} & & 
\multicolumn{2}{c}{R@100 = 99.2} & & 
\multicolumn{2}{c}{R@100 = 99.5} & & 
\multicolumn{2}{c}{R@100 = 99.8} & & 
\multicolumn{2}{c}{R@100 = 99.9} & & 
\multicolumn{2}{c}{R@100 = 100.0} & & &
\multicolumn{2}{c}{R@100 = 98.8} \\
\cline{2-3} \cline{5-6} \cline{8-9} \cline{11-12} \cline{14-15} \cline{17-18} \cline{20-21} \cline{23-24} \cline{26-27} \cline{29-30} \cline{33-34}
& R@1 & R@10 & & R@1 & R@10 & & R@1 & R@10 & & R@1 & R@10 & & R@1 & R@10 & & R@1 & R@10 & & R@1 & R@10 & & R@1  & R@10 & & R@1  & R@10 & & R@1 & R@10 & & & R@1 & R@10 \\
\hline
- & 94.9 & 99.8 && \textbf{95.6} & \textbf{97.6} && \textbf{98.7} & \textbf{99.9} && 74.2 & 84.3 && 72.4 & 86.8 && \textbf{96.4} & 98.2 && \underline{98.7} & \textbf{99.5} && \underline{97.7} & \underline{99.4} && \underline{98.7} & \textbf{99.8} && 97.5 & \underline{99.7} &&& \underline{92.5} & 96.5 \\
\hline
R2D2 (NeurIPS '19) & 95.9 & \underline{99.9} && 80.9 & 91.8 && 95.5 & 99.7 && 52.1 & 76.2 && 64.5 & 85.5 && 89.0 & 95.7 && 97.5 & \underline{99.3} && 76.7 & 91.9 && 90.0 & 97.2 && 91.1 & 98.4 &&& 83.3 & 93.6 \\
D2Net (CVPR '19) & 96.6 & \textbf{100.0} && 84.2 & 94.1 && 95.0 & 99.7 && 68.5 & 82.8 && 72.4 & 90.8 && 92.7 & 97.3 && 98.2 & \textbf{99.5} && 84.6 & 96.8 && 92.2 & 98.4 && 95.6 & 99.0 &&& 88.0 & 95.8 \\
SuperGlue (CVPR '20) & 97.1 & \textbf{100.0} && 91.9 & \underline{96.6} && 96.8 & \underline{99.8} && \textbf{78.1} & \underline{86.5} && \underline{80.3} & \textbf{93.4} && 95.9 & \underline{98.6} && 97.8 & \textbf{99.5} && 91.4 & 98.5 && 94.7 & 99.2 && 96.8 & 99.4 &&& 92.1 & \textbf{97.2} \\
LoFTR (CVPR '21) & 97.7 & \textbf{100.0} && 91.0 & 96.4 && 96.7 & \underline{99.8} && 75.3 & 85.4 && \underline{80.3} & 90.8 && 95.0 & 97.8 && 98.3 & \textbf{99.5} && 95.5 & 98.9 && 97.7 & 99.3 && 97.5 & \textbf{100.0} &&& \underline{92.5} & \underline{96.8} \\
Patch2Pix (CVPR '21) & 96.3 & \underline{99.9} && 84.8 & 93.9 && 96.1 & 99.7 && 54.7 & 76.0 && 75.0 & 90.8 && 92.0 & 96.9 && 97.0 & 99.2 && 70.8 & 93.9 && 94.8 & 98.9 && 95.6 & 98.7 &&& 85.7 & 94.8 \\
Matchformer (ACCV '22) & 97.4 & \textbf{100.0} && 88.7 & 94.9 && 97.4 & \underline{99.8} && \textbf{78.1} & 85.6 && 78.9 & 88.2 && 95.9 & 98.1 && \textbf{98.8} & \textbf{99.5} && 94.8 & 99.3 && 97.7 & \underline{99.4} && 97.1 & \underline{99.7} &&& \underline{92.5} & 96.5 \\
SuperPoint+LightGlue (ICCV '23) & 97.2 & \textbf{100.0} && 90.1 & 96.1 && 97.0 & \underline{99.8} && \underline{76.6} & \underline{86.5} && \underline{80.3} & 90.8 && \underline{96.2} & \underline{98.6} && 97.7 & \underline{99.3} && 93.7 & 98.1 && 97.0 & \underline{99.4} && \underline{98.7} & 99.4 &&& \underline{92.5} & \underline{96.8} \\
DISK+LightGlue (ICCV '23) & 97.0 & \textbf{100.0} && 91.9 & 96.2 && 95.9 & 99.7 && 72.3 & 83.5 && 77.6 & \underline{92.1} && 94.0 & 98.2 && 97.8 & \textbf{99.5} && 84.0 & 92.3 && 94.6 & 98.1 && 96.2 & \underline{99.7} &&& 90.1 & 95.9 \\
ALIKED+LightGlue (ICCV '23) & \textbf{98.1} & \textbf{100.0} && \underline{93.6} & \underline{96.6} && 97.3 & 99.7 && 73.2 & 83.9 && \underline{80.3} & \underline{92.1} && 95.7 & \underline{98.6} && 98.5 & \textbf{99.5} && 86.1 & 94.5 && 91.2 & 97.2 && 98.4 & \underline{99.7} &&& 91.2 & 96.2 \\
RoMa (CVPR '24) & 95.4 & 99.8 && 58.4 & 92.5 && 96.1 & \underline{99.8} && 64.2 & \textbf{87.8} && 67.1 & 90.8 && 91.8 & 97.7 && 97.8 & \textbf{99.5} && 83.8 & 99.1 && 78.0 & 97.5 && 96.5 & \underline{99.7} &&& 82.9 & 96.4 \\
Tiny-RoMa (CVPR '24) & 95.4 & 99.8 && 83.8 & 94.7 && 95.9 & 99.7 && 60.5 & 82.0 && 65.8 & 89.5 && 91.8 & 97.7 && 97.8 & \textbf{99.5} && 58.8 & 93.4 && 78.0 & 97.5 && 90.5 & 99.0 &&& 81.8 & 95.3 \\
Steerers (CVPR '24) & 97.3 & \underline{99.9} && 82.7 & 91.3 && 97.3 & \textbf{99.9} && 67.8 & 81.5 && 72.4 & 88.2 && 94.3 & 98.0 && 98.5 & \textbf{99.5} && 87.0 & 96.7 && 96.1 & 98.6 && \underline{98.7} & \textbf{100.0} &&& 89.2 & 95.4 \\
Affine Steerers (ECCV '24) & 95.7 & 99.7 && 85.6 & 93.9 && 97.3 & \textbf{99.9} && 69.7 & 82.4 && 72.4 & 88.2 && 94.2 & 98.1 && 98.5 & \textbf{99.5} && 87.0 & 96.7 && 95.3 & 98.5 && 97.1 & 99.4 &&& 89.3 & 95.6 \\
DUSt3R (CVPR '24) & 94.2 & \textbf{100.0} && 68.8 & 86.0 && 94.4 & 99.5 && 54.9 & 71.9 && 67.1 & 84.2 && 88.4 & 94.9 && 97.2 & \underline{99.3} && 67.7 & 80.3 && 84.2 & 91.3 && 92.1 & 96.8 &&& 80.9 & 90.4 \\
MASt3R (ECCV '24) & 96.4 & \textbf{100.0} && 81.5 & 96.0 && \underline{98.4} & \textbf{99.9} && 75.5 & 85.6 && \underline{80.3} & \textbf{93.4} && 95.9 & \underline{98.6} && 97.8 & \textbf{99.5} && \textbf{98.2} & \textbf{99.6} && \textbf{99.2} & \textbf{99.8} && \textbf{99.7} & \textbf{100.0} &&& 92.3 & \textbf{97.2} \\
xFeat (CVPR '24) & 94.0 & \underline{99.9} && 88.3 & 94.9 && 95.9 & 99.7 && 61.2 & 80.9 && 63.2 & 90.8 && 90.7 & 97.0 && 97.7 & \textbf{99.5} && 80.4 & 94.2 && 87.6 & 97.3 && 92.4 & \underline{99.7} &&& 85.1 & 95.4 \\
GIM-DKMv3 (ICLR '24) & 75.6 & 99.7 && 9.1 & 48.6 && 87.0 & \underline{99.8} && 58.2 & 83.3 && 53.9 & 89.5 && 46.6 & 92.2 && 73.9 & 98.8 && 65.2 & 97.1 && 60.5 & 98.9 && 76.5 & \underline{99.7} &&& 60.6 & 90.8 \\
GIM-LightGlue (ICLR '24) & \underline{97.9} & \textbf{100.0} && 92.0 & \underline{96.6} && 97.3 & \textbf{99.9} && 76.0 & 84.8 && \textbf{81.6} & 89.5 && \textbf{96.4} & \textbf{98.8} && 98.0 & \textbf{99.5} && 93.6 & 97.8 && 97.0 & \underline{99.4} && 98.4 & \underline{99.7} &&& \textbf{92.8} & 96.6 \\
\bottomrule
\end{tabular}
\end{adjustbox}
\vspace{-3mm}
\caption{\textbf{Recalls before and after applying re-ranking, with a threshold of 100 meters.}
The shortlist of candidates to be re-ranked is obtained with MegaLoc, and the results with such shortlist are shown in the first row.
Re-ranking has been applied to the first 100 candidates (\ie $K=100$). Next to each dataset's name, we show the R@100, which in practice sets the upper bound of the maximum recalls achievable after re-ranking. Best results are in \textbf{bold}, second best are \underline{underlined}.
}
\label{tab:reranking_results_megaloc_100m}
\vspace{-3mm}
\end{table*}

\myparagraph{Results:} We articulate our analysis on the relationship between prediction uncertainty, and re-ranking performance through \cref{fig:histogram_prob}, \cref{tab:auprc_megaloc_25m}, and \cref{tab:reranking_results_megaloc_25m}.
In \cref{tab:auprc_megaloc_25m}
we present the results of uncertainty estimation, across multiple datasets, of existing baselines for uncertainty estimation applied directly on MegaLoc predictions, and several matching methods, for which we use the number of inliers as a confidence score.
In \cref{fig:histogram_prob}, we train a Logistic Regressor to predict the probability of a query being a correct match based on the number of inliers on the top-1 prediction. We train the Logistic Regressor on MASt3R inliers counts on MSLS val, and plot the resulting histogram of probabilities for Pitts30k and SF-XL Occlusion.

From this data, we draw the following conclusions:
\begin{itemize}
    \item \textbf{Low Uncertainty translates in re-ranking being detrimental}. On datasets where MegaLoc achieves 95+\% R@1, applying re-ranking worsens performances. In this scenario, there is little  uncertainty on retrieval predictions (\textit{cf}. \cref{fig:histogram_prob});
    \item \textbf{High Uncertainty leaves room for improvement via re-ranking}: on Baidu, SF-XL Night and Occlusion, uncertainty is higher (\textit{cf}. \cref{fig:histogram_prob}), and on these datasets re-ranking generally improves R@1. For instance, MASt3R provides a boost of respectively +2.1\%, +5.4\%, +5.3\%;
    \item \textbf{Image Matching methods are better at estimating uncertainty }. On saturated datasets, even a \textit{Random} uncertainty estimator achieves an AUPRC of over 90\% (\textit{cf}. \cref{tab:auprc_megaloc_25m}). On the other hand, on the challenging SF-XL Night and Occlusion, using inlier count as a proxy of uncertainty is consistently better than existing baselines in terms of AUC (\textit{cf}. \cref{tab:reranking_results_megaloc_25m}).
\end{itemize}

\myparagraph{Additional insights.} In contrast with previous literature \cite{Zaffar_2024_estimation}, we see that L2-distance (in MegaLoc's feature space) can be a fairly good estimator of uncertainty: we hypothesize this discrepancy to be due to MegaLoc being a more robust model w.r.t. VPR models analyzed in previous uncertainty estimation papers \cite{Zaffar_2024_estimation, Warburg_2021_bayesian, Cai_2022_stun, Hausler_2021_unsupervised}.

Lastly, \cref{fig:precision_recall_curves_megaloc} shows the PR curves for Tokyo 24/7 and two of the most challenging datasets for MegaLoc, namely SF-XL Night and Occlusion. These curves illustrate how (i) retrieval alone achieves the ideal curve on saturated benchmarks such as Tokyo; and (ii)
image matching methods significantly enhance uncertainty estimation, in scenarios in which the retrieval model actually struggles.

\subsection{Additional Experiments}
\label{sec:ablation}

\myparagraph{Effect of the distance threshold}
\label{sec:re_ranking_100m}
A potential question is whether our observation that re-ranking can degrade performance is solely due to the 25-meter threshold. It’s possible that image matching methods have been trained to recognize broader views of the same location, potentially placing images slightly beyond the 25-meter threshold among the top predictions after re-ranking. To investigate this, we recompute the results using a 100-meter threshold to determine if this is indeed the case.
The results on re-ranking for VPR with $\tau$ set to 100 meters are presented in \cref{tab:reranking_results_megaloc_100m}.
The table shows that our findings are indeed robust to the choice of $\tau$, as it can be seen that re-ranking can have a negative impact on results on 4 datasets.

\begin{figure*}
\centering
\begin{subfigure}{0.49\linewidth}
    \begin{subfigure}{0.24\linewidth}
      \centering
      \includegraphics[width=\linewidth]{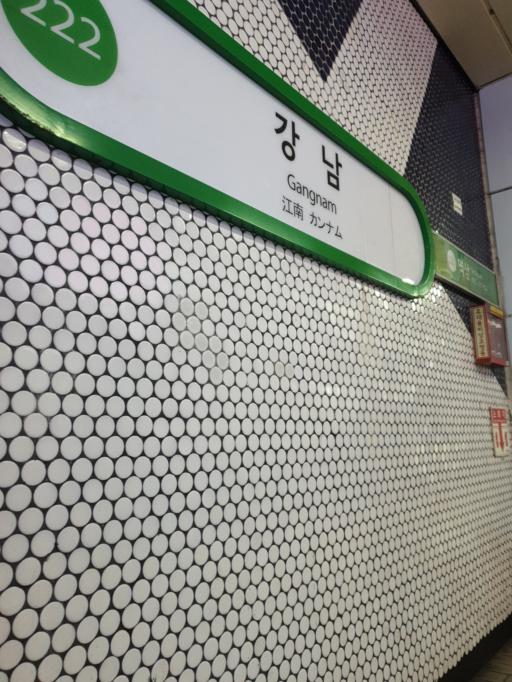}
      \label{fig:query_baidu}
    \end{subfigure}%
    \hfill\begin{subfigure}{0.24\linewidth}
      \centering
      \includegraphics[width=\linewidth]{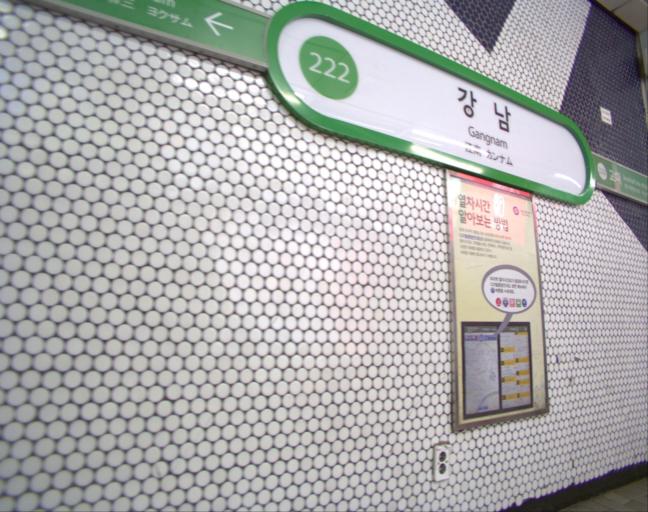}
      \label{fig:db_baidu}
    \end{subfigure}%
    \hfill\begin{subfigure}{0.5\linewidth}
      \centering
      \includegraphics[width=\linewidth]{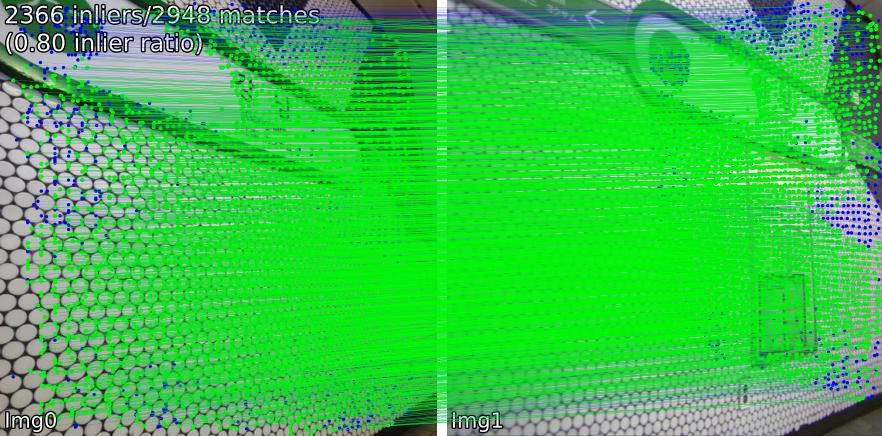}
      \label{fig:match_baidu}
    \end{subfigure}
     \vspace{-1.3em} 
    \caption{Baidu -- inliers: 2366, distance: 119m}
  \label{fig:baidu_wrong_query}
\end{subfigure}%
\hfill\begin{subfigure}{0.49\linewidth}
    \begin{subfigure}{0.24\linewidth}
      \centering
      \includegraphics[width=\linewidth]{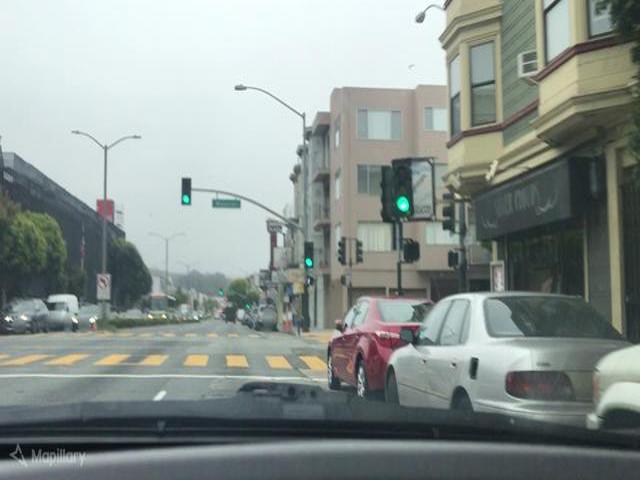}
      \label{fig:query_msls_val}
    \end{subfigure}%
    \hfill\begin{subfigure}{0.24\linewidth}
      \centering
      \includegraphics[width=\linewidth]{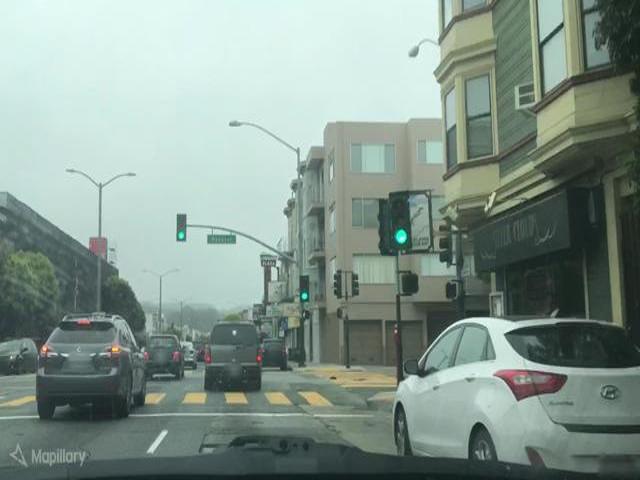}
      \label{fig:db_msls_val}
    \end{subfigure}%
    \hfill\begin{subfigure}{0.5\linewidth}
      \centering
      \includegraphics[width=\linewidth]{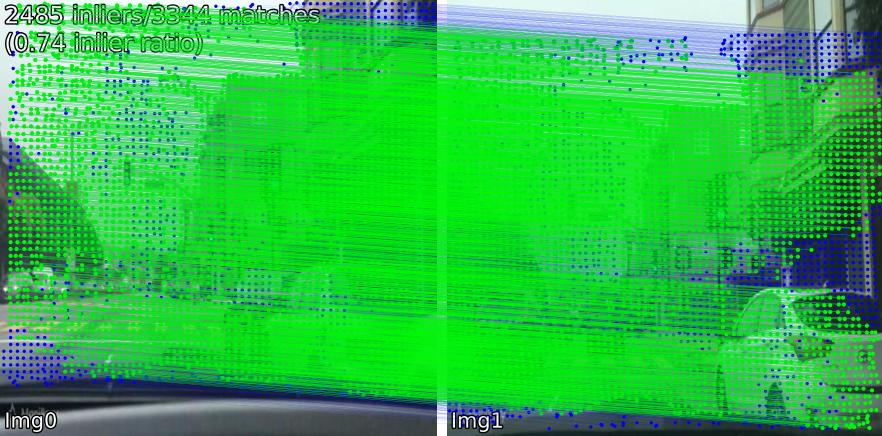}
      \label{fig:match_msls_val}
    \end{subfigure}
     \vspace{-1.3em} 
    \caption{MSLS Val -- inliers: 2485, distance: 182m}
  \label{fig:msls_wrong_query}
\end{subfigure}

\vspace{1em} 

\begin{subfigure}{0.49\linewidth}
  \begin{subfigure}{0.24\linewidth}
    \centering
    \includegraphics[width=\linewidth]{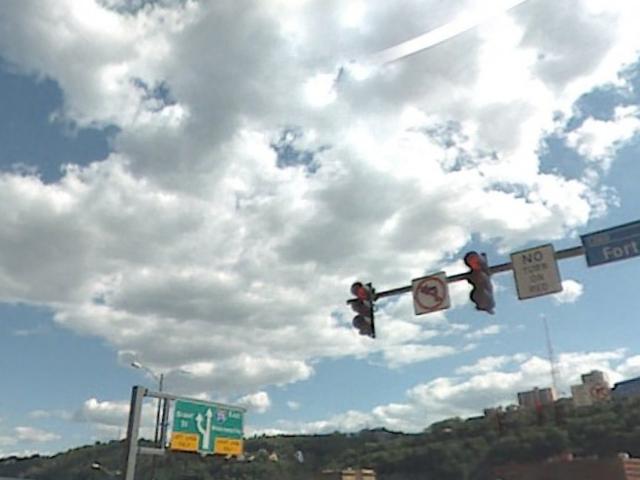}
    \label{fig:query_pitts30k}
  \end{subfigure}%
  \hfill\begin{subfigure}{0.24\linewidth}
    \centering
    \includegraphics[width=\linewidth]{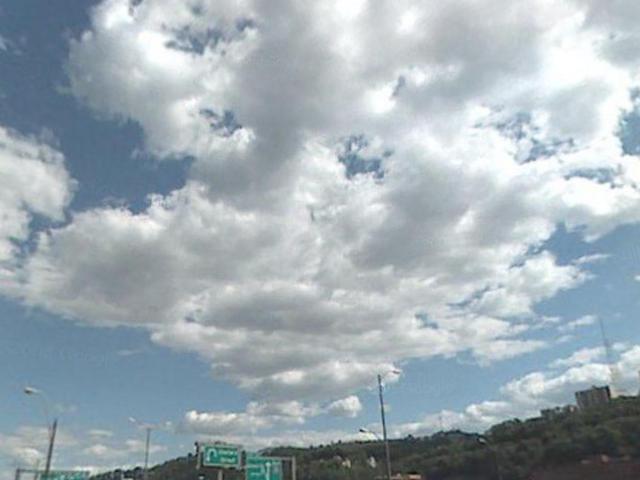}
    \label{fig:db_pitts30k}
  \end{subfigure}%
  \hfill\begin{subfigure}{0.5\linewidth}
    \centering
    \includegraphics[width=\linewidth]{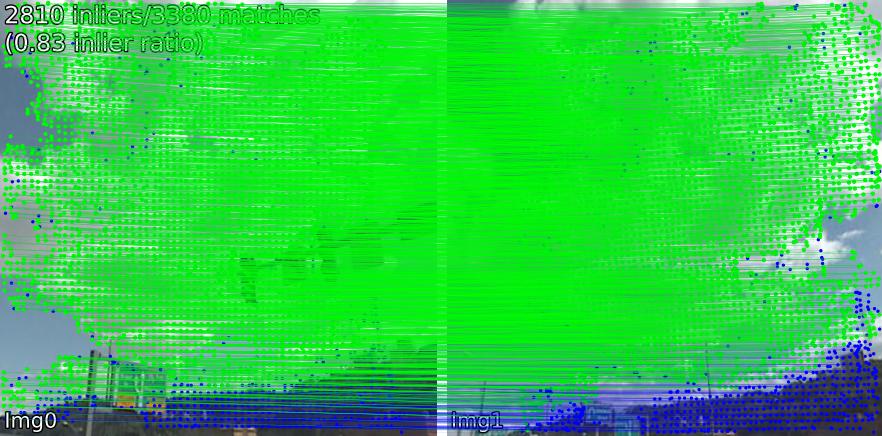}
    \label{fig:match_pitts30k}
  \end{subfigure}
     \vspace{-1.3em} 
    \caption{Pitts30k -- inliers: 2810, distance: 60m}
  \label{fig:pitts30k_wrong_query}
\end{subfigure}%
\hfill\begin{subfigure}{0.49\linewidth}
  \begin{subfigure}{0.24\linewidth}
    \centering
    \includegraphics[width=\linewidth]{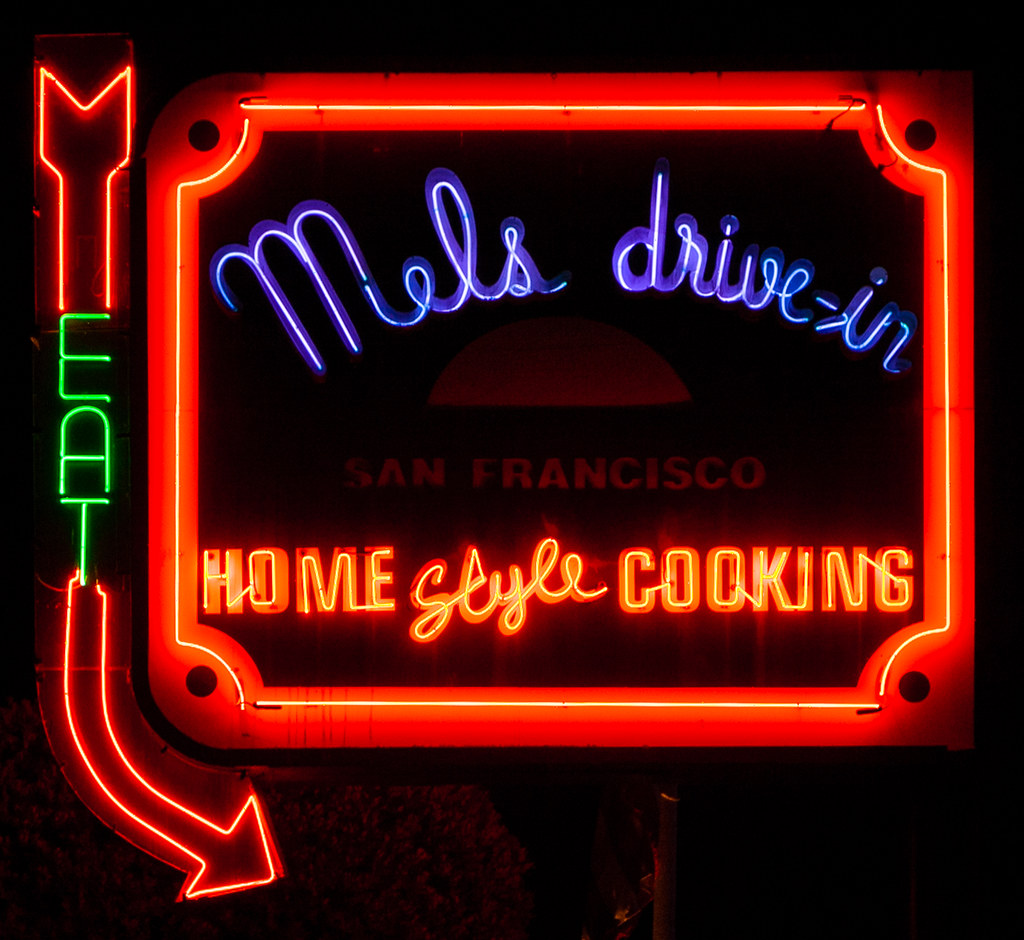}
    \label{fig:query_sf_xl_night}
  \end{subfigure}%
  \hfill\begin{subfigure}{0.24\linewidth}
    \centering
    \includegraphics[width=\linewidth]{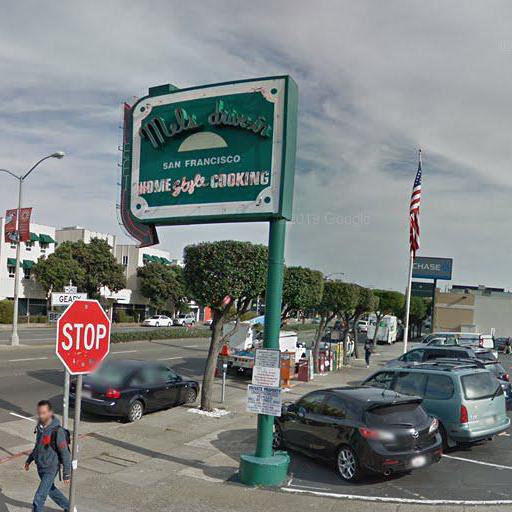}
    \label{fig:db_sf_xl_night}
  \end{subfigure}%
  \hfill\begin{subfigure}{0.5\linewidth}
    \centering
    \includegraphics[width=\linewidth]{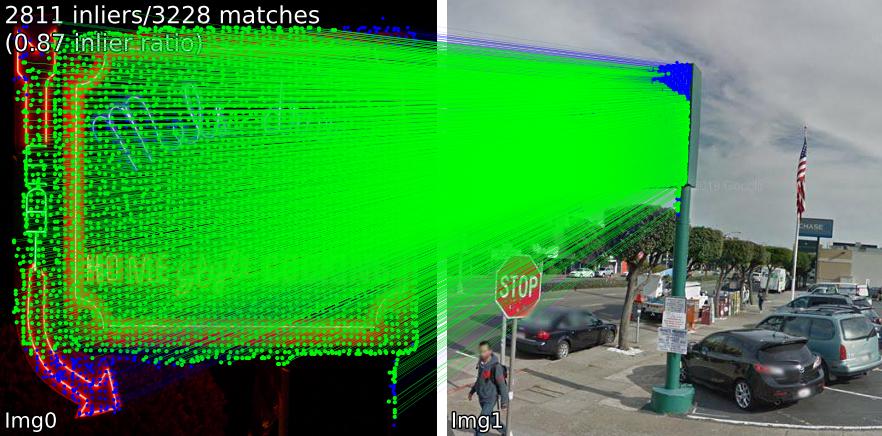}
    \label{fig:match_sf_xl_night}
  \end{subfigure}
     \vspace{-1.3em} 
  \caption{SF-XL Night -- inliers: 2811, distance: 29m}
  \label{fig:sf_xl_night_wrong_query}
\end{subfigure}

\vspace{1em} 

\begin{subfigure}{0.49\linewidth}
  \begin{subfigure}{0.24\linewidth}
    \centering
    \includegraphics[width=\linewidth]{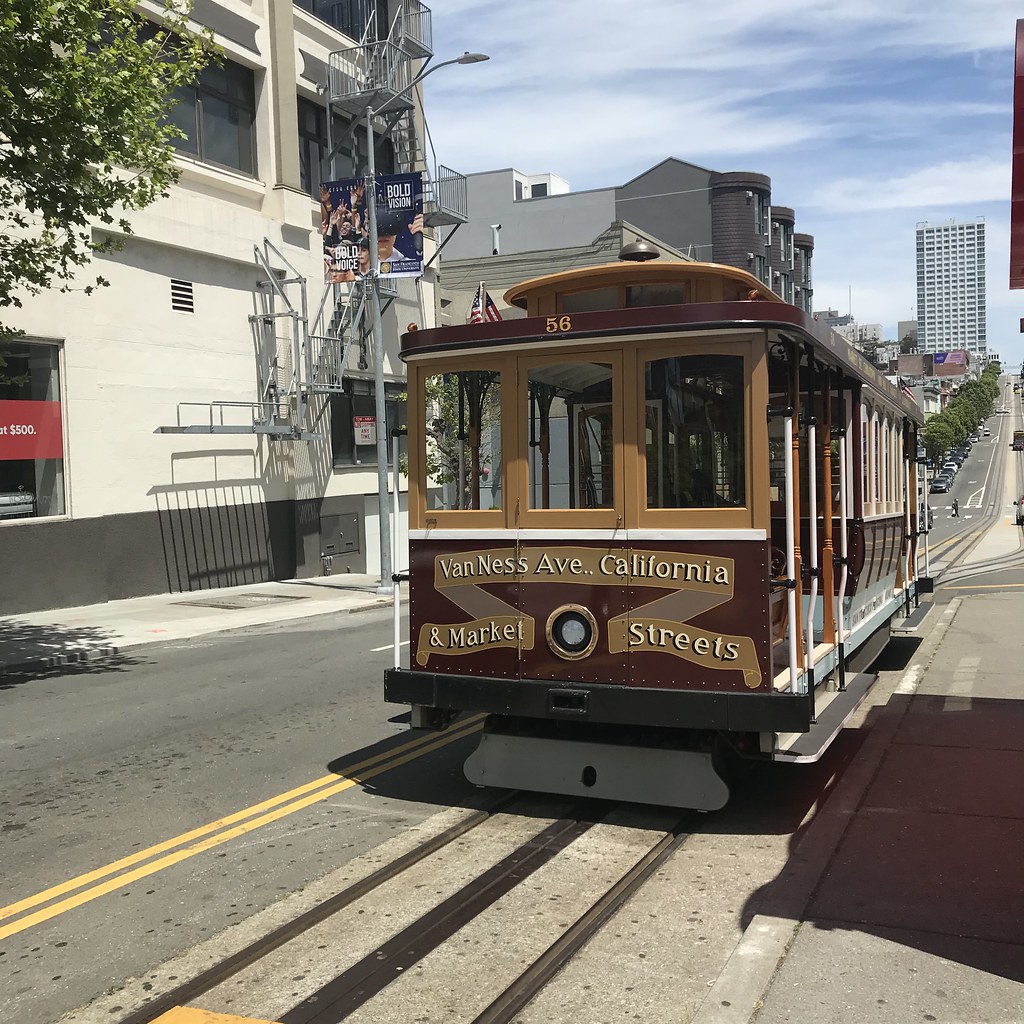}
    \label{fig:query_sf_xl_occlusion}
  \end{subfigure}%
  \hfill\begin{subfigure}{0.24\linewidth}
    \centering
    \includegraphics[width=\linewidth]{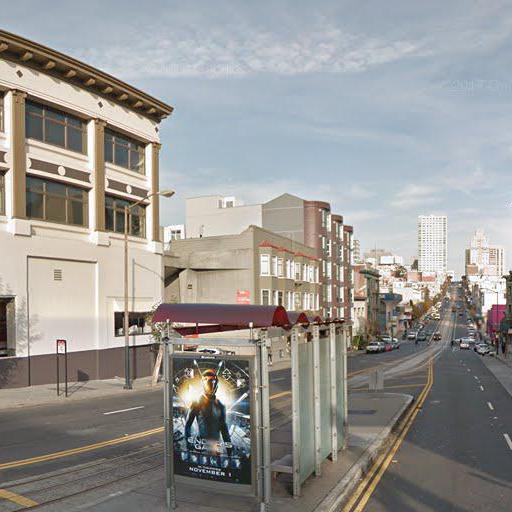}
    \label{fig:db_sf_xl_occlusion}
  \end{subfigure}%
  \hfill\begin{subfigure}{0.5\linewidth}
    \centering
    \includegraphics[width=\linewidth]{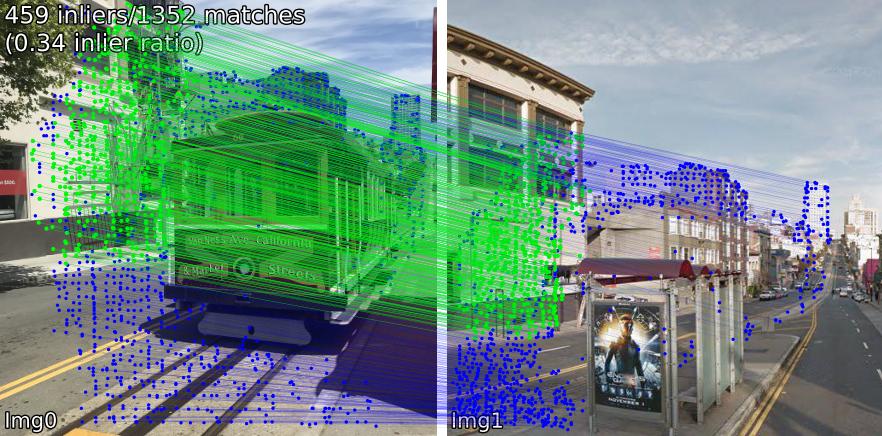}
    \label{fig:match_sf_xl_occlusion}
  \end{subfigure}
     \vspace{-1.3em} 
  \caption{SF-XL Occlusion -- inliers: 459, distance: 34m}
  \label{fig:sf_xl_occlusion_wrong_query}
\end{subfigure}%
\hfill\begin{subfigure}{0.49\linewidth}
  \begin{subfigure}{0.24\linewidth}
    \centering
    \includegraphics[width=\linewidth]{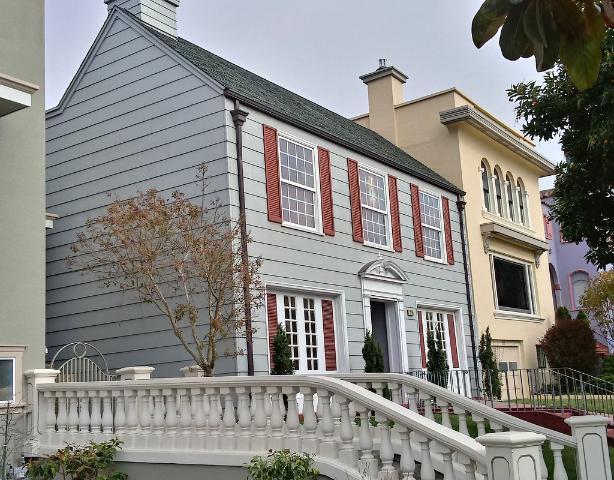}
    \label{fig:query_sf_xl_v1}
  \end{subfigure}%
  \hfill\begin{subfigure}{0.24\linewidth}
    \centering
    \includegraphics[width=\linewidth]{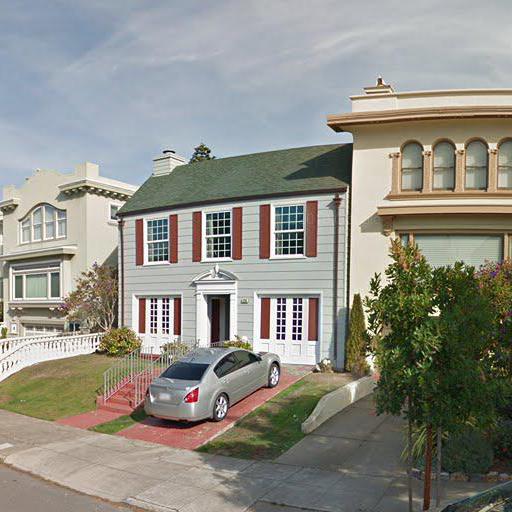}
    \label{fig:db_sf_xl_v1}
  \end{subfigure}%
  \hfill\begin{subfigure}{0.5\linewidth}
    \centering
    \includegraphics[width=\linewidth]{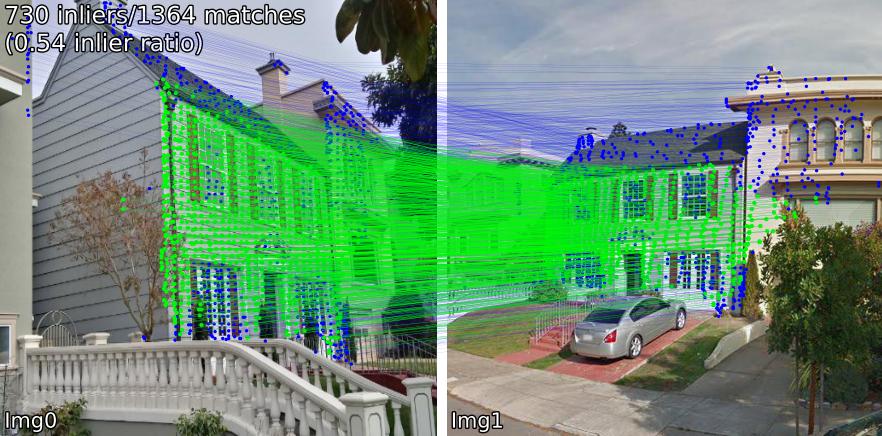}
    \label{fig:match_sf_xl_v1}
  \end{subfigure}
     \vspace{-1.3em} 
  \caption{SF-XL test V1 -- inliers: 730, distance: 26m}
  \label{fig:sf_xl_v1_wrong_query}
\end{subfigure}

\vspace{1em} 

\begin{subfigure}{0.49\linewidth}
  \begin{subfigure}{0.24\linewidth}
    \centering
    \includegraphics[width=\linewidth]{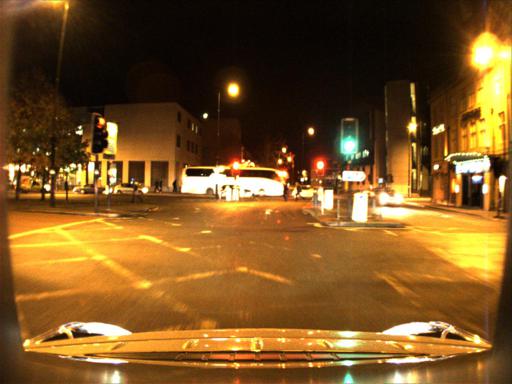}
    \label{fig:query_svox_night}
  \end{subfigure}%
  \hfill\begin{subfigure}{0.24\linewidth}
    \centering
    \includegraphics[width=\linewidth]{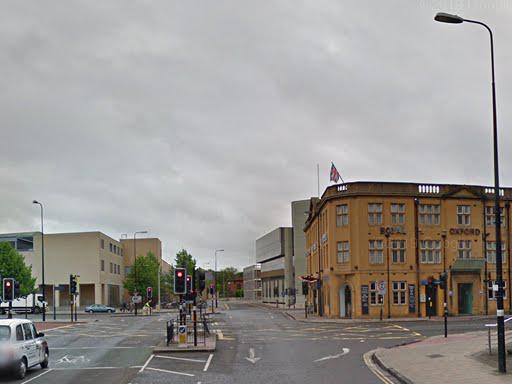}
    \label{fig:db_svox_night}
  \end{subfigure}%
  \hfill\begin{subfigure}{0.5\linewidth}
    \centering
    \includegraphics[width=\linewidth]{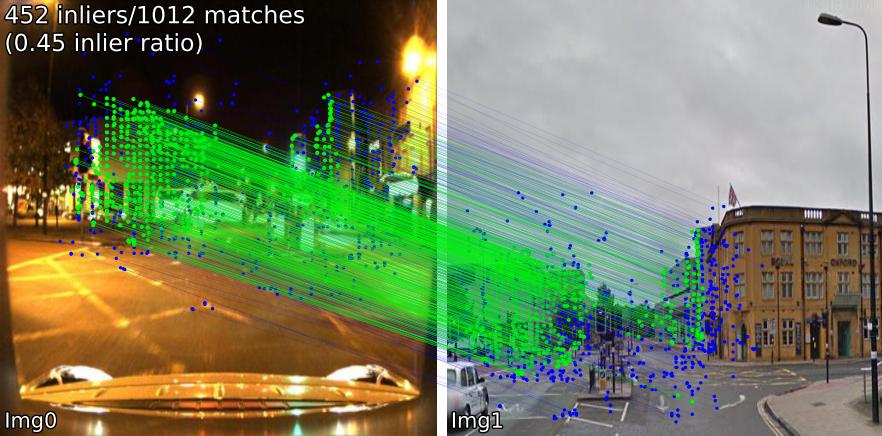}
    \label{fig:match_svox_night}
  \end{subfigure}
     \vspace{-1.3em} 
  \caption{SVOX Night -- inliers: 452, distance: 27m}
  \label{fig:svox_night_wrong_query}
\end{subfigure}%
\hfill\begin{subfigure}{0.49\linewidth}
  \begin{subfigure}{0.24\linewidth}
    \centering
    \includegraphics[width=\linewidth]{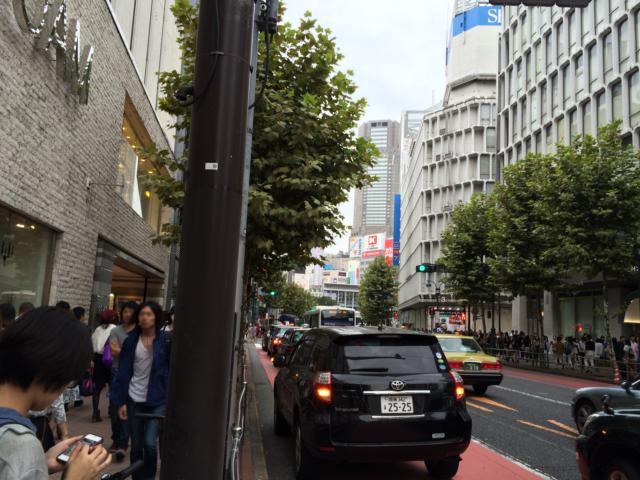}
    \label{fig:query_tokyo247}
  \end{subfigure}%
  \hfill\begin{subfigure}{0.24\linewidth}
    \centering
    \includegraphics[width=\linewidth]{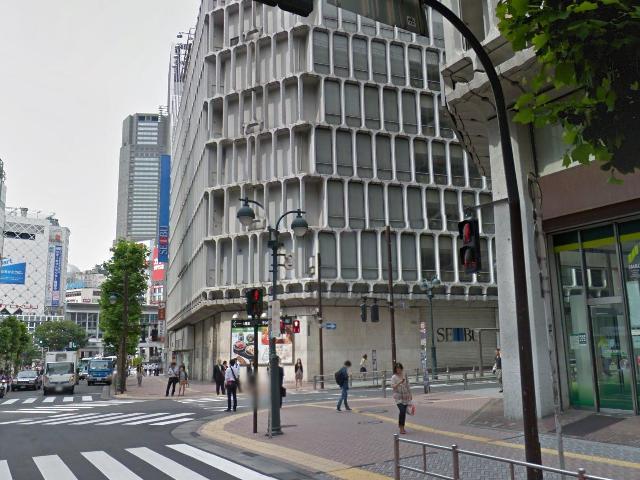}
    \label{fig:db_tokyo247}
  \end{subfigure}%
  \hfill\begin{subfigure}{0.5\linewidth}
    \centering
    \includegraphics[width=\linewidth]{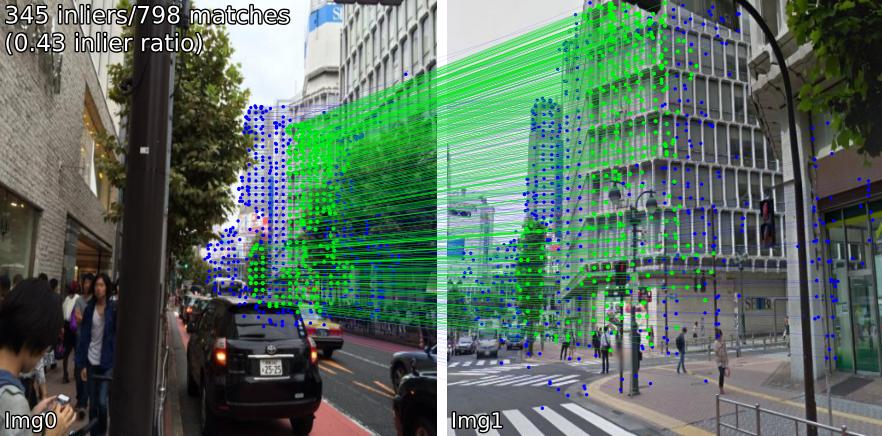}
    \label{fig:match_tokyo247}
  \end{subfigure}
     \vspace{-1.3em} 
  \caption{Tokyo 24/7 -- inliers: 345, distance: 35m}
  \label{fig:tokyo247_wrong_query}
\end{subfigure}
\vspace{-2mm}
\caption{\textbf{Wrong queries with the largest number of inliers.} For each dataset, the displayed images show the query and a confidently retrieved negative image. The dataset name is accompanied by the number of inliers and the distance (in meters) between the two images, as indicated by their labels.}
\label{fig:matches_all_datasets}
\vspace{-4mm}
\end{figure*}

\subsubsection{Failure cases with largest number of inliers}
\label{sec:failures_with_high_number_of_inliers}

In this section, we provide a deeper analysis of the types of failures encountered by image matching methods in uncertainty estimation. We focus on MASt3R as the representative method, given its consistency across all experiments and datasets and its recent development. Additional examples for the remaining datasets are reported in the Supplementary Material.
For each dataset, we identify the query with the highest number of inliers, but where the nearest neighbor retrieved by MegaLoc is located more than 25 meters away from the query's ground-truth location, effectively leading to a wrong (but confident) prediction.
These cases are illustrated in ~\cref{fig:matches_all_datasets}, which shows the queries, the top-1 retrieved database images, the matched inliers, and the distance in meters between each pair of images. Based on this, we categorize the failures into two distinct types, as described below.

\textbf{Noisy GPS Labels / Perceptual Aliasing.}
The first category of failure arises from pairs of images that look to be from the same place but, according to the GPS labels, are from different places.
This is clear in the examples from MSLS, Baidu, SF-XL occlusion.
While some of these cases might be due to perceptual aliasing, \ie the phenomenon for which two different places look almost identical (which can happen especially in indoor places, like in the case of Baidu), we believe that in many cases this is due to noisy GPS coordinates: it should be noted that GPS labels, even when post-processed (\eg Mapillary famously post-processes images' GPS with SfM \cite{Warburg_2020_msls}), can be wrong.
As examples, it can be noted the pairs of images from MSLS, which, although the distance according to the GPS is over 50 meters, it is likely that the two photos have been taken from a much smaller distance.

\textbf{Large Distance and Viewpoint Variation.} The second category consists of failure cases where the retrieved image is just above the 25 meters threshold, as in the cases of Tokyo 24/7 and SF-XL V1, whose predictions from \cref{fig:matches_all_datasets} are 35 and 26 meters away.

\subsubsection{Limitations}
\label{sec:limitations}

This paper presents, among other insights, the most comprehensive benchmark for re-ranking in VPR, both in terms of models and datasets.
Although we aimed to obtain results that are as fair and comparable to each other, it must be noted that some of the chosen hyperparameters could benefit one method over another: for example, the image resolution was set to $512 \times 512$, which is a common resolution in VPR \cite{Berton_2022_cosPlace, Alibey_2022_gsvcities}, and some image matching methods might benefit from this more than others (\eg RoMa \cite{Edstedt_2024_roma} is known to prefer higher resolutions).

\section{Conclusions}
\label{sec:conclusions}

In this work, we revisit the conventional retrieval-and-re-ranking pipeline in the context of recent advances in the field. Our findings reveal that current state-of-the-art retrieval methods have effectively saturated historically challenging benchmarks, uncovering a counter-intuitive side effect: re-ranking can degrade performance when applied to near-perfect predictions. The key insight of this work is that image matching methods remain valuable; not as a default mechanism for trading computational cost for performance, but as a strategic tool to assess the confidence of retrieval predictions. When uncertainty is detected, image matching can then be employed selectively to refine and improve results. Finally, we present a comprehensive benchmark of re-ranking techniques for visual place recognition, encompassing a diverse range of methods and datasets to guide future research in the field.

\myparagraph{Acknowledgements}
\noindent Gabriele Trivigno and Carlo Masone were supported by FAIR - Future Artificial Intelligence Research which received funding from the European Union Next-GenerationEU (PIANO NAZIONALE DI RIPRESA E RESILIENZA (PNRR) – MISSIONE 4 COMPONENTE 2, INVESTIMENTO 1.3 – D.D. 1555 11/10/2022, PE00000013).  This manuscript reflects only the authors’ views and opinions, neither the European Union nor the European Commission can be considered responsible for them.

\noindent We acknowledge the CINECA award
under the ISCRA initiative, for the availability of high performance computing resources.
{
    \small
    \bibliographystyle{ieeenat_fullname}
    \bibliography{main}
}

\clearpage
\setcounter{page}{1}
\maketitlesupplementary
\appendix
\section{Additional Results}
\begin{figure*}
\centering

\begin{subfigure}{0.49\linewidth}
  \begin{subfigure}{0.24\linewidth}
    \centering
    \includegraphics[width=\linewidth]{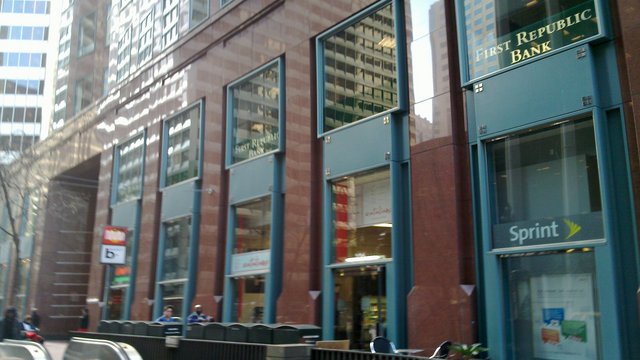}
    \label{fig:query_sf_xl_v2}
  \end{subfigure}%
  \hfill\begin{subfigure}{0.24\linewidth}
    \centering
    \includegraphics[width=\linewidth]{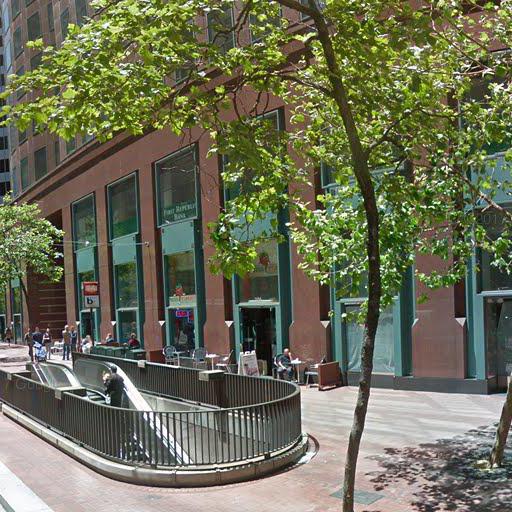}
    \label{fig:db_sf_xl_v2}
  \end{subfigure}%
  \hfill\begin{subfigure}{0.5\linewidth}
    \centering
    \includegraphics[width=\linewidth]{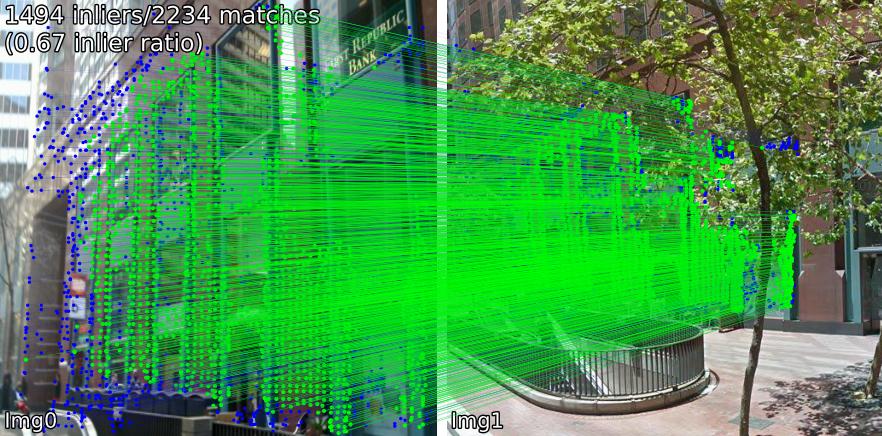}
    \label{fig:match_sf_xl_v2}
  \end{subfigure}
     \vspace{-1.2em} 
  \caption{SF-XL test V2 -- inliers: 1494, distance: 79m}
  \label{fig:sf_xl_v2_wrong_query}
\end{subfigure}%
\hfill\begin{subfigure}{0.49\linewidth}
  \begin{subfigure}{0.24\linewidth}
    \centering
    \includegraphics[width=\linewidth]{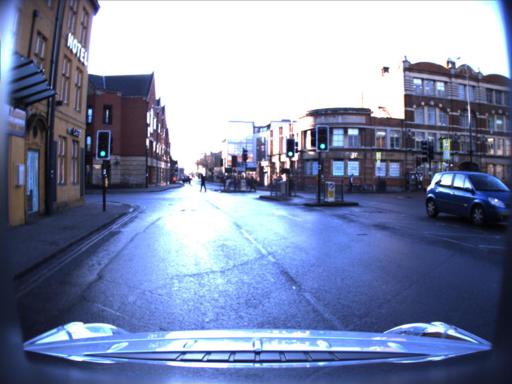}
    \label{fig:query_svox_sun}
  \end{subfigure}%
  \hfill\begin{subfigure}{0.24\linewidth}
    \centering
    \includegraphics[width=\linewidth]{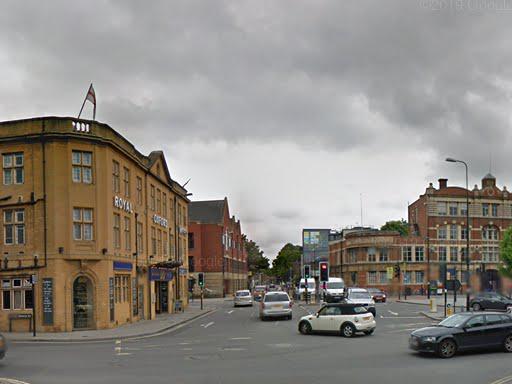}
    \label{fig:db_svox_sun}
  \end{subfigure}%
  \hfill\begin{subfigure}{0.5\linewidth}
    \centering
    \includegraphics[width=\linewidth]{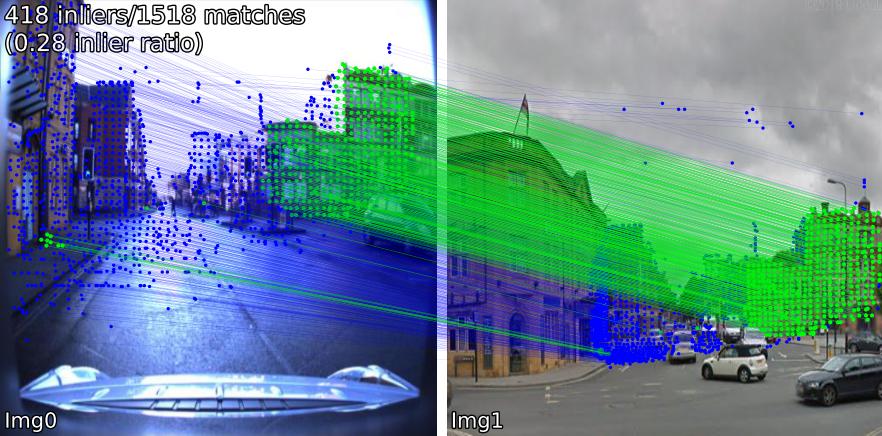}
    \label{fig:match_svox_sun}
  \end{subfigure}
     \vspace{-1.2em} 
  \caption{SVOX Sun -- inliers: 418, distance: 29m}
  \label{fig:svox_sun_wrong_query}
\end{subfigure}

\vspace{1em} 

\vspace{-2mm}
\caption{\textbf{Wrong queries with the largest number of inliers for SF-XL test V2 and SVOX Sun.} The displayed images show the query and a confidently retrieved negative image. The dataset name is accompanied by the number of inliers and the distance (in meters) between the two images, as indicated by their labels.}
\label{fig:matches_remaining_datasets_supp}
\vspace{-4mm}
\end{figure*}

\begin{figure*}
\centering
\begin{subfigure}{0.49\linewidth}
  \centering
  \includegraphics[width=0.7\linewidth]{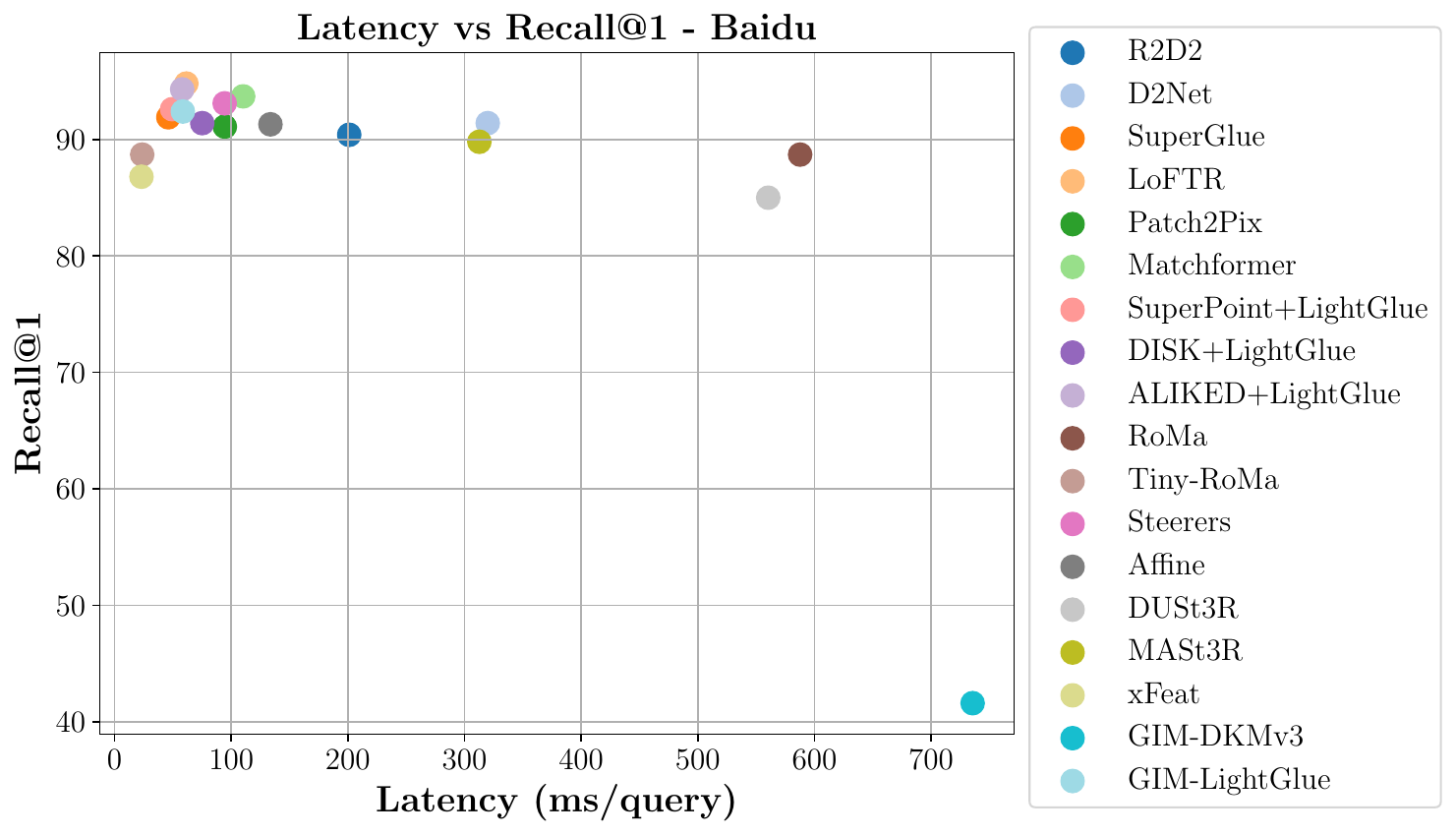}
  \label{fig:latency_baidu}
\end{subfigure}%
\hfill\begin{subfigure}{0.49\linewidth}
  \centering
  \includegraphics[width=0.7\linewidth]{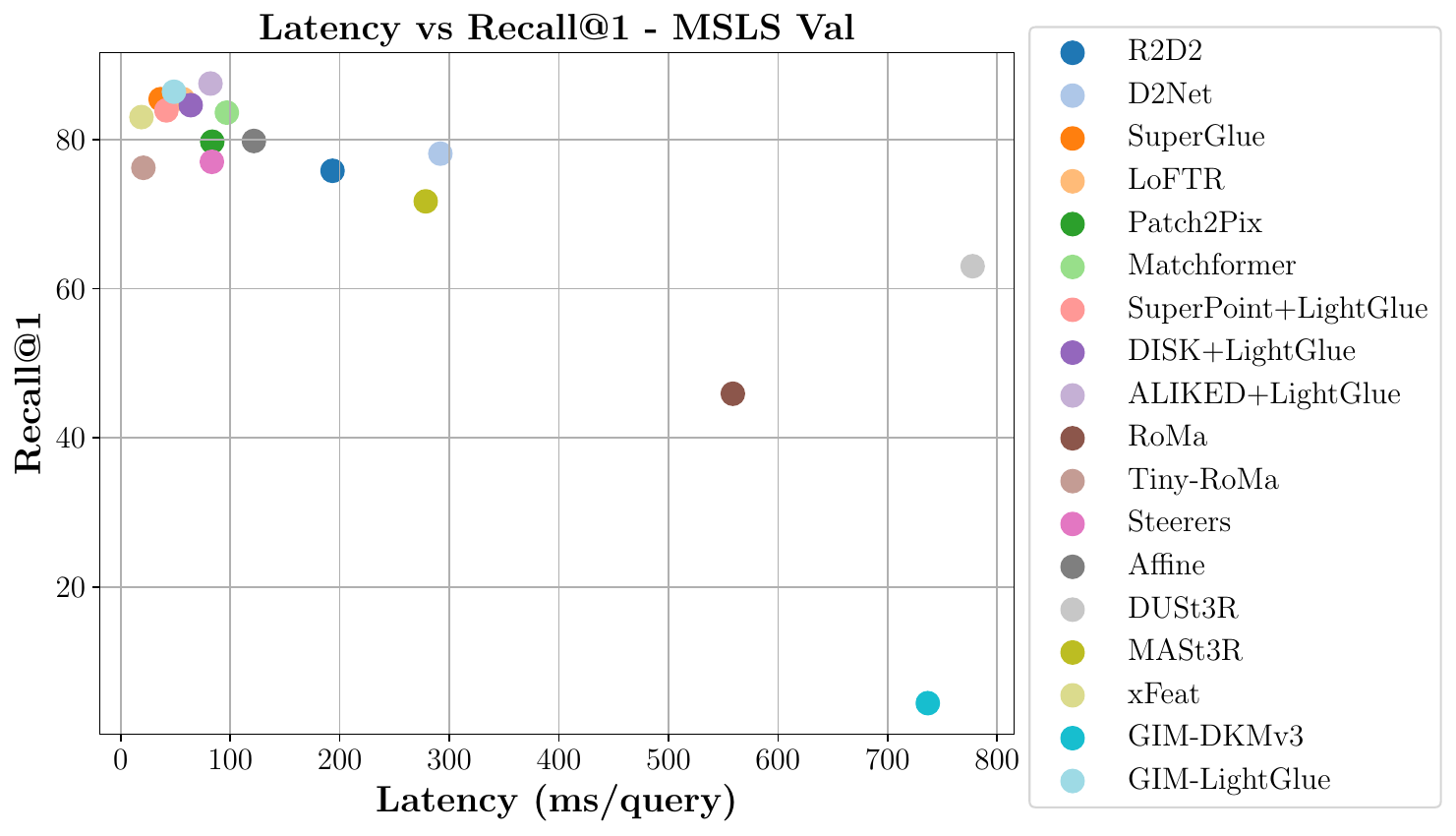}
  \label{fig:latency_msls}
\end{subfigure}

\begin{subfigure}{0.49\linewidth}
  \centering
  \includegraphics[width=0.7\linewidth]{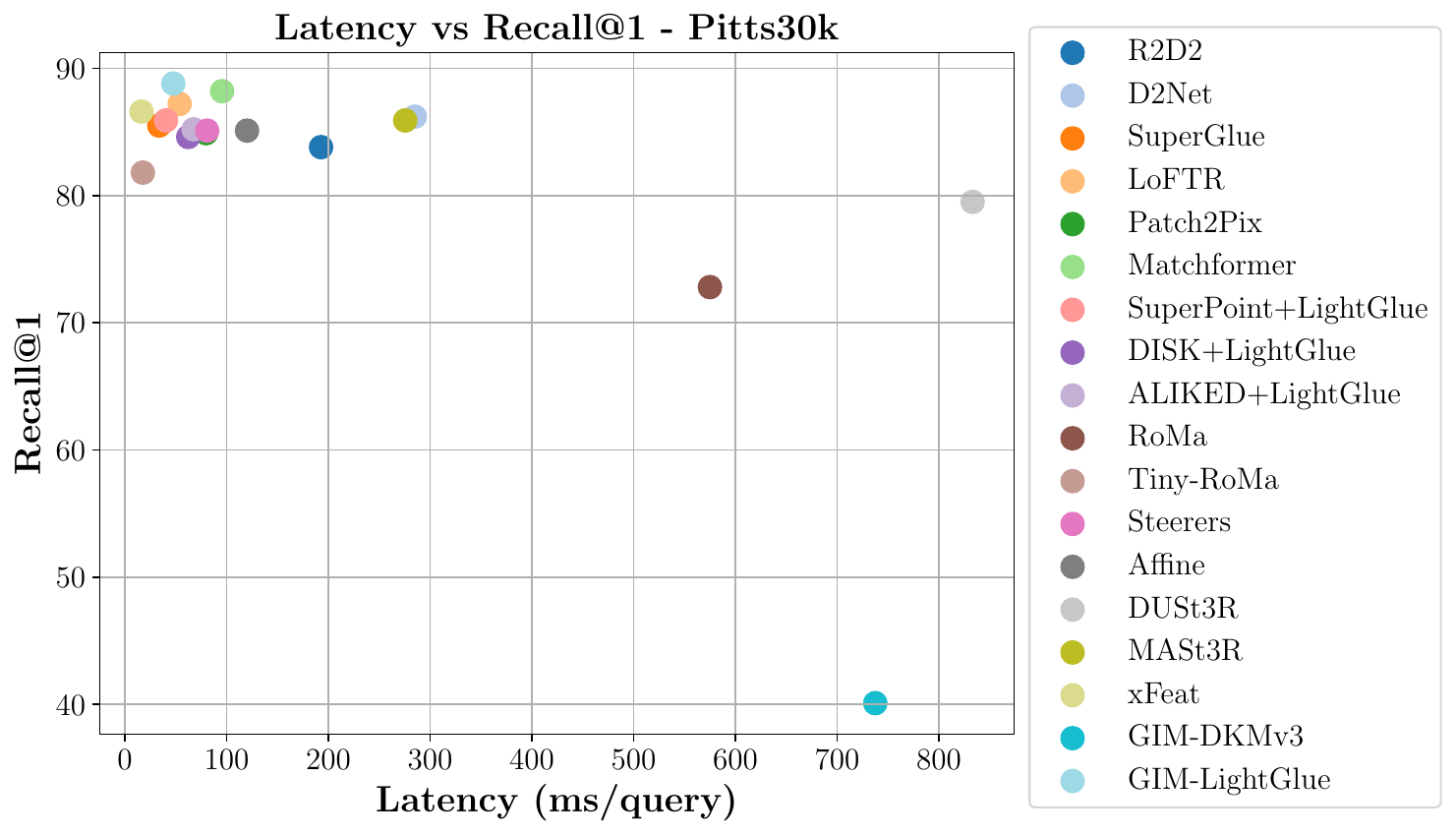}
  \label{fig:latency_pitts30k}
\end{subfigure}%
\hfill\begin{subfigure}{0.49\linewidth}
  \centering
  \includegraphics[width=0.7\linewidth]{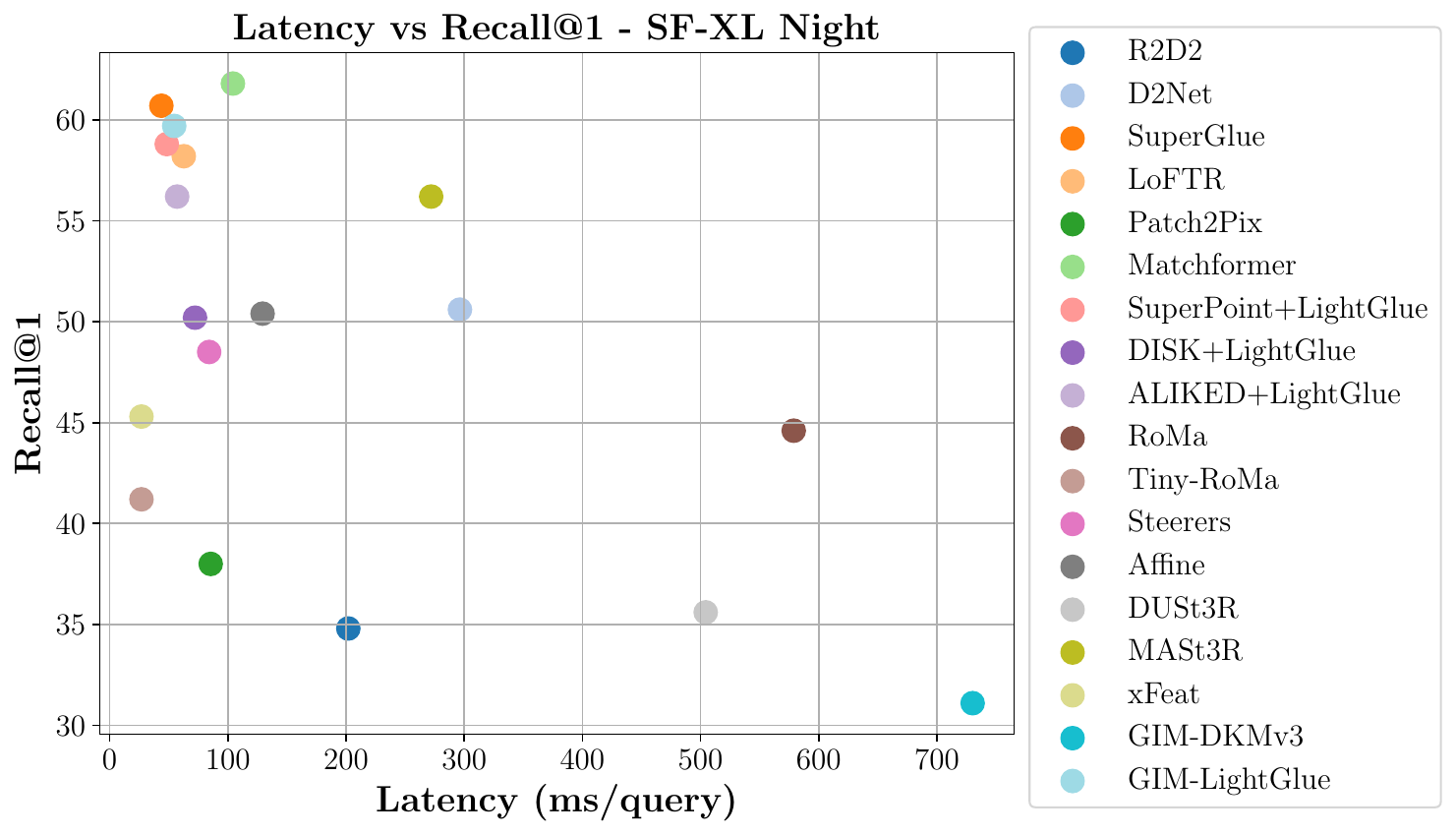}
  \label{fig:latency_sf_xl_night}
\end{subfigure}

\begin{subfigure}{0.49\linewidth}
  \centering
  \includegraphics[width=0.7\linewidth]{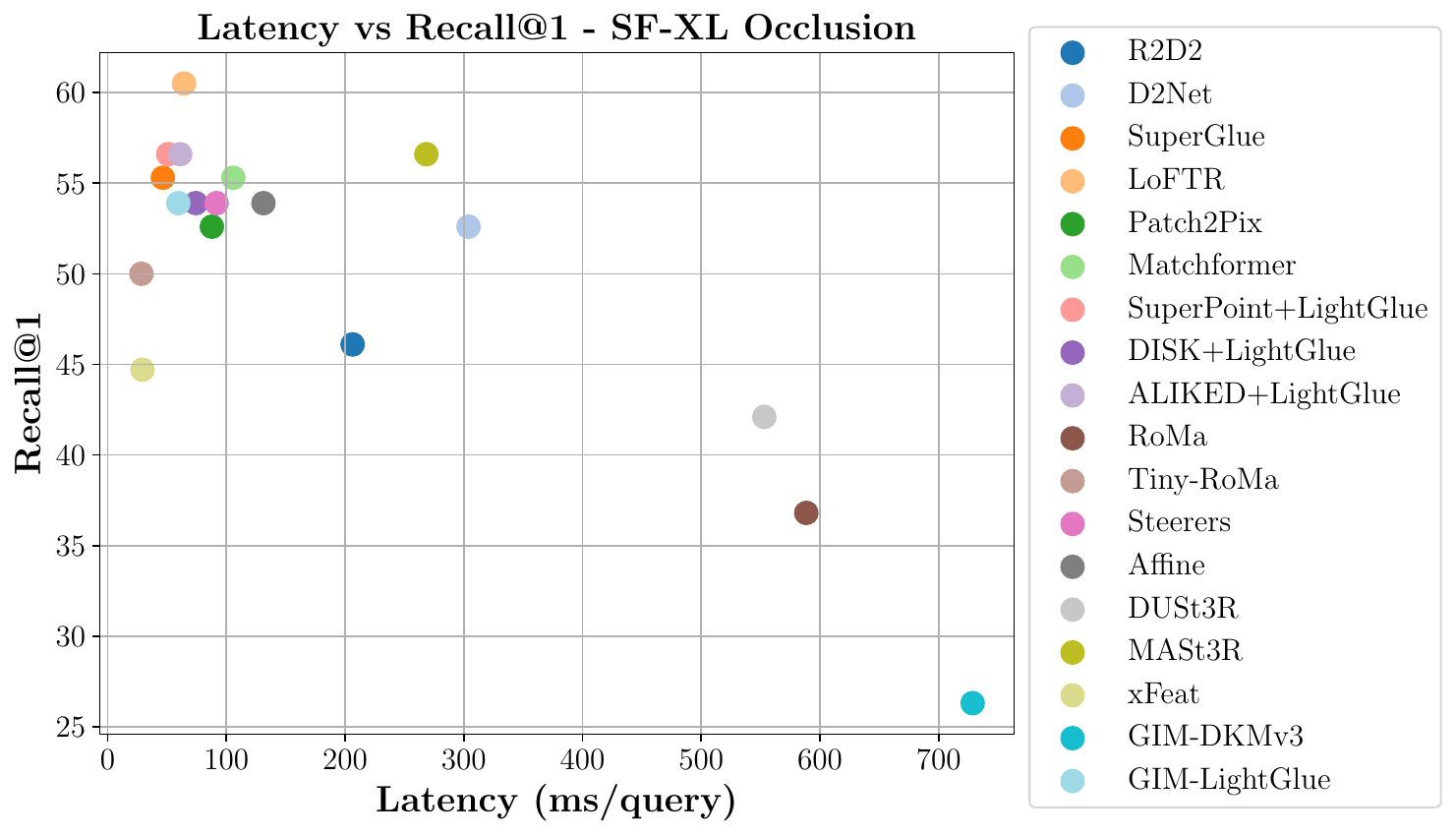}
  \label{fig:latency_sf_xl_occlusion}
\end{subfigure}%
\hfill\begin{subfigure}{0.49\linewidth}
  \centering
  \includegraphics[width=0.7\linewidth]{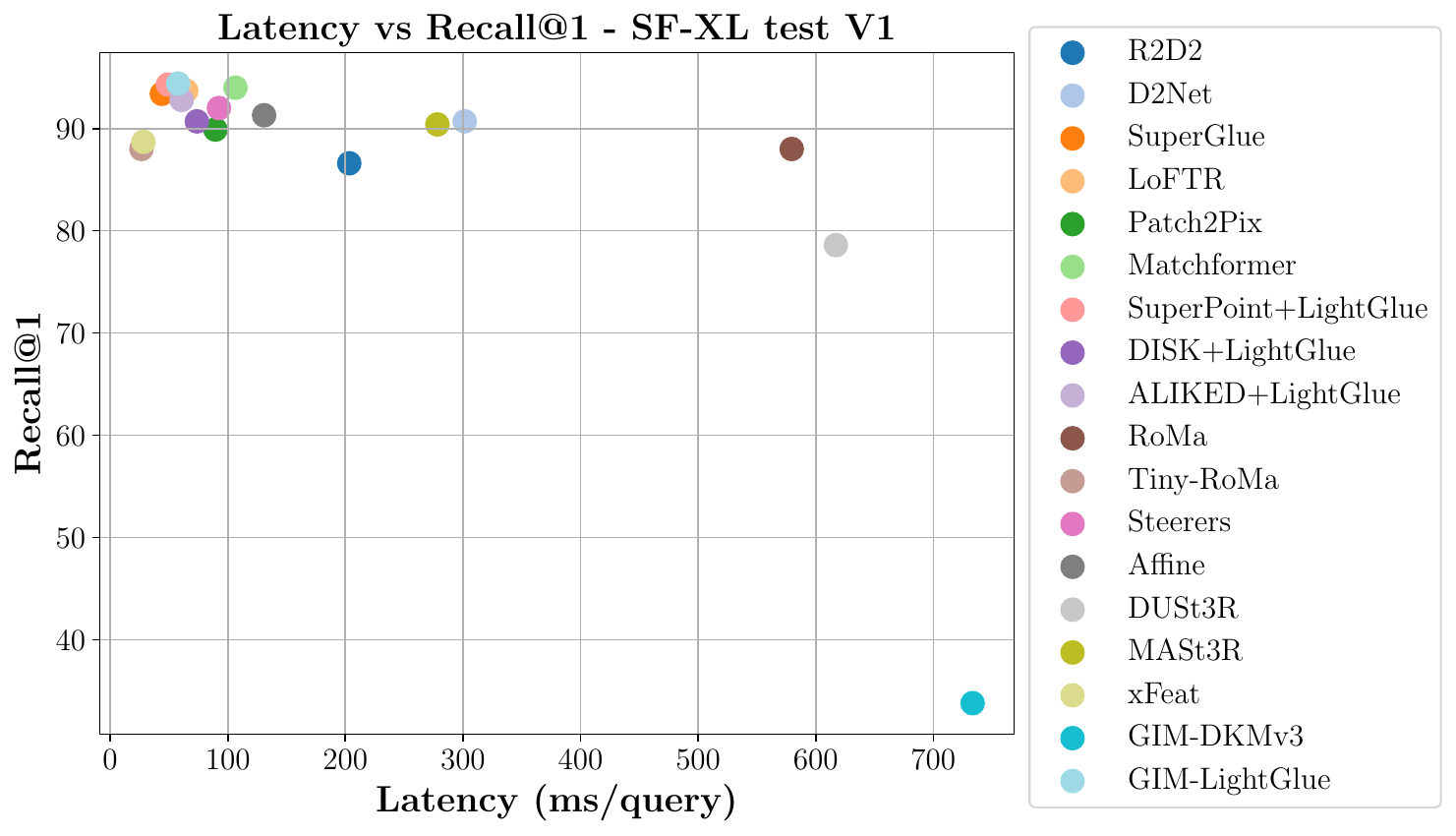}
  \label{fig:latency_sf_xl_v1}
\end{subfigure}

\begin{subfigure}{0.49\linewidth}
  \centering
  \includegraphics[width=0.7\linewidth]{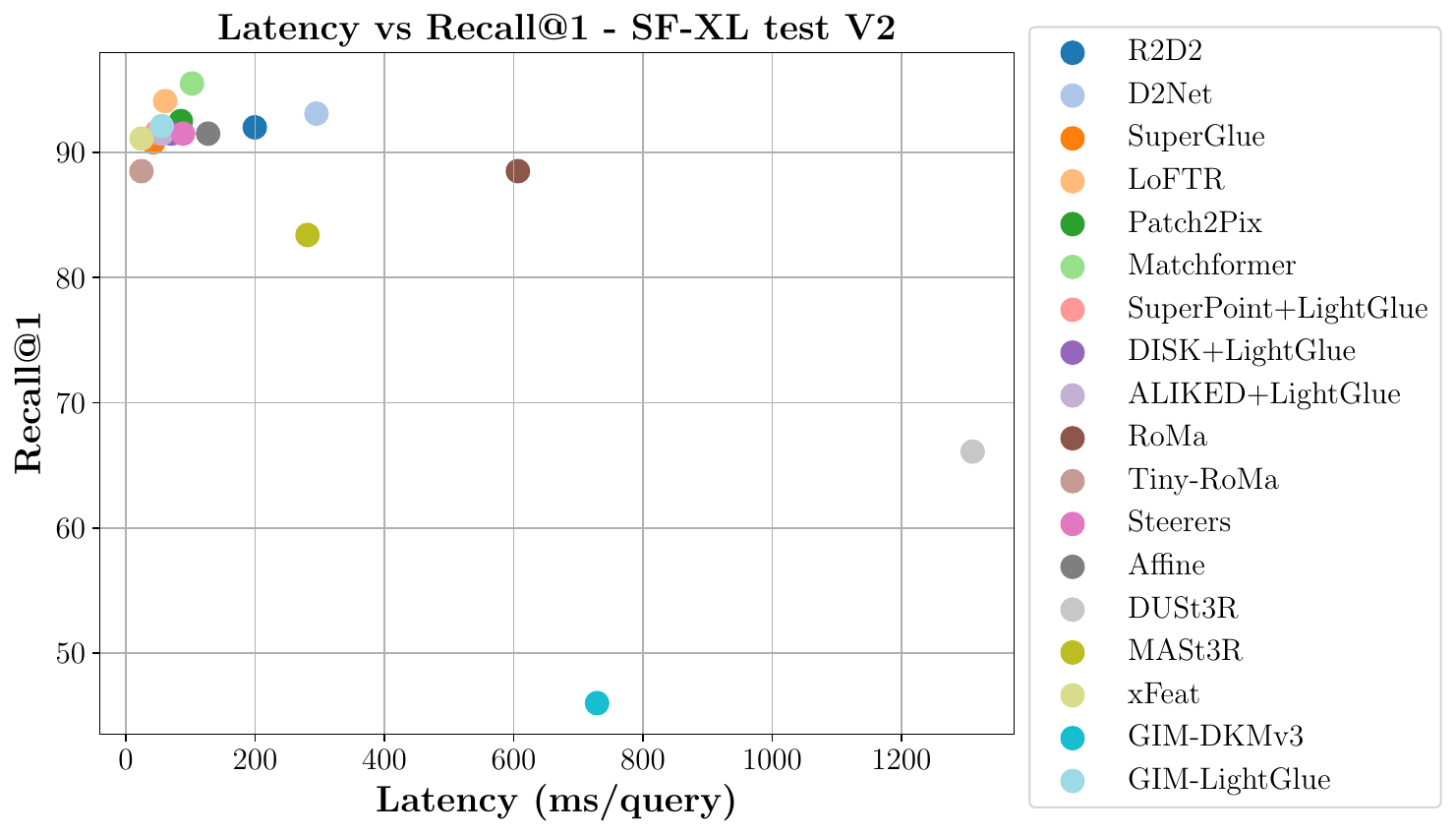}
  \label{fig:latency_sf_xl_v2}
\end{subfigure}%
\hfill\begin{subfigure}{0.49\linewidth}
  \centering
  \includegraphics[width=0.7\linewidth]{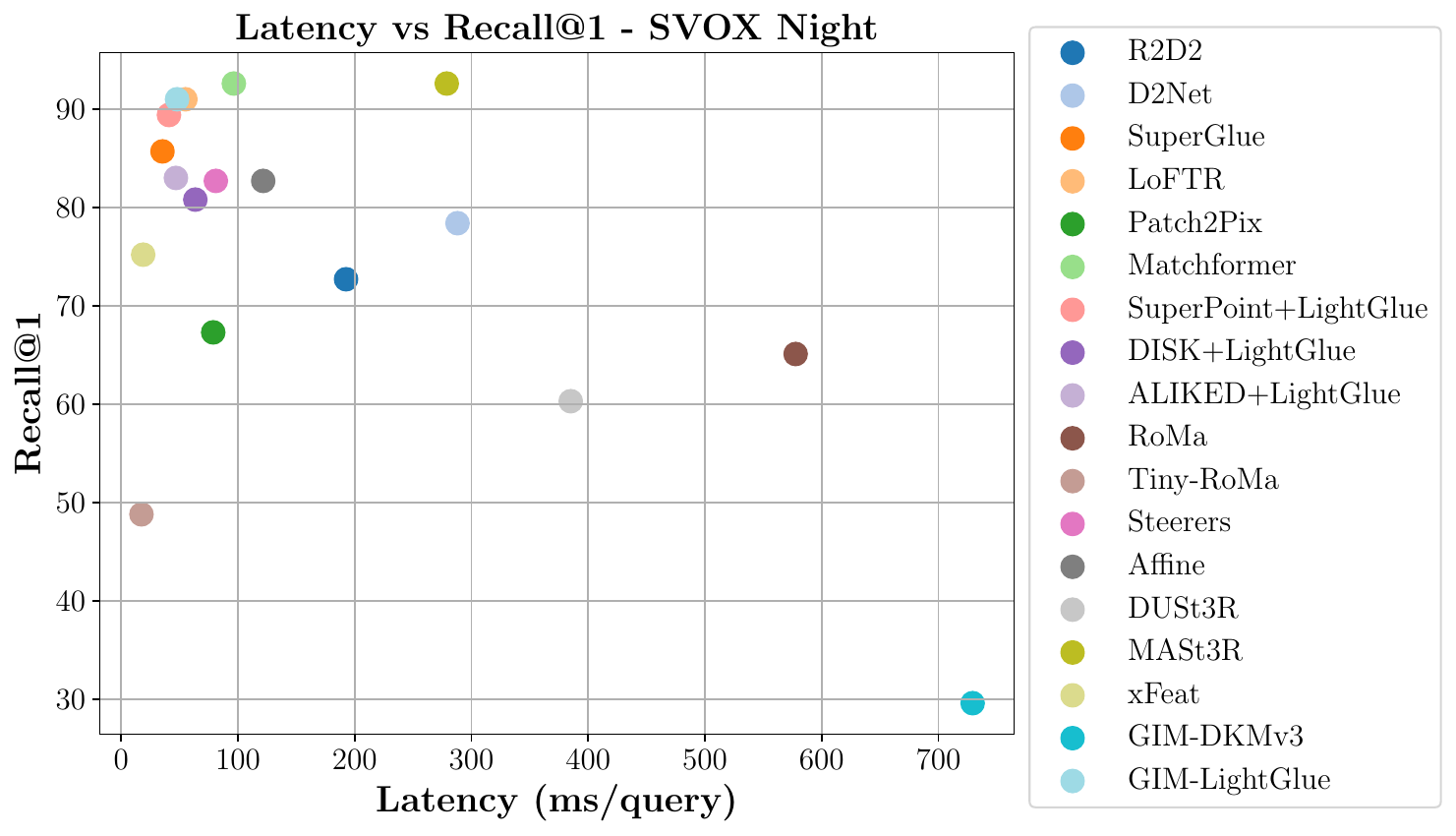}
  \label{fig:latency_svox_night}
\end{subfigure}

\begin{subfigure}{0.49\linewidth}
  \centering
  \includegraphics[width=0.7\linewidth]{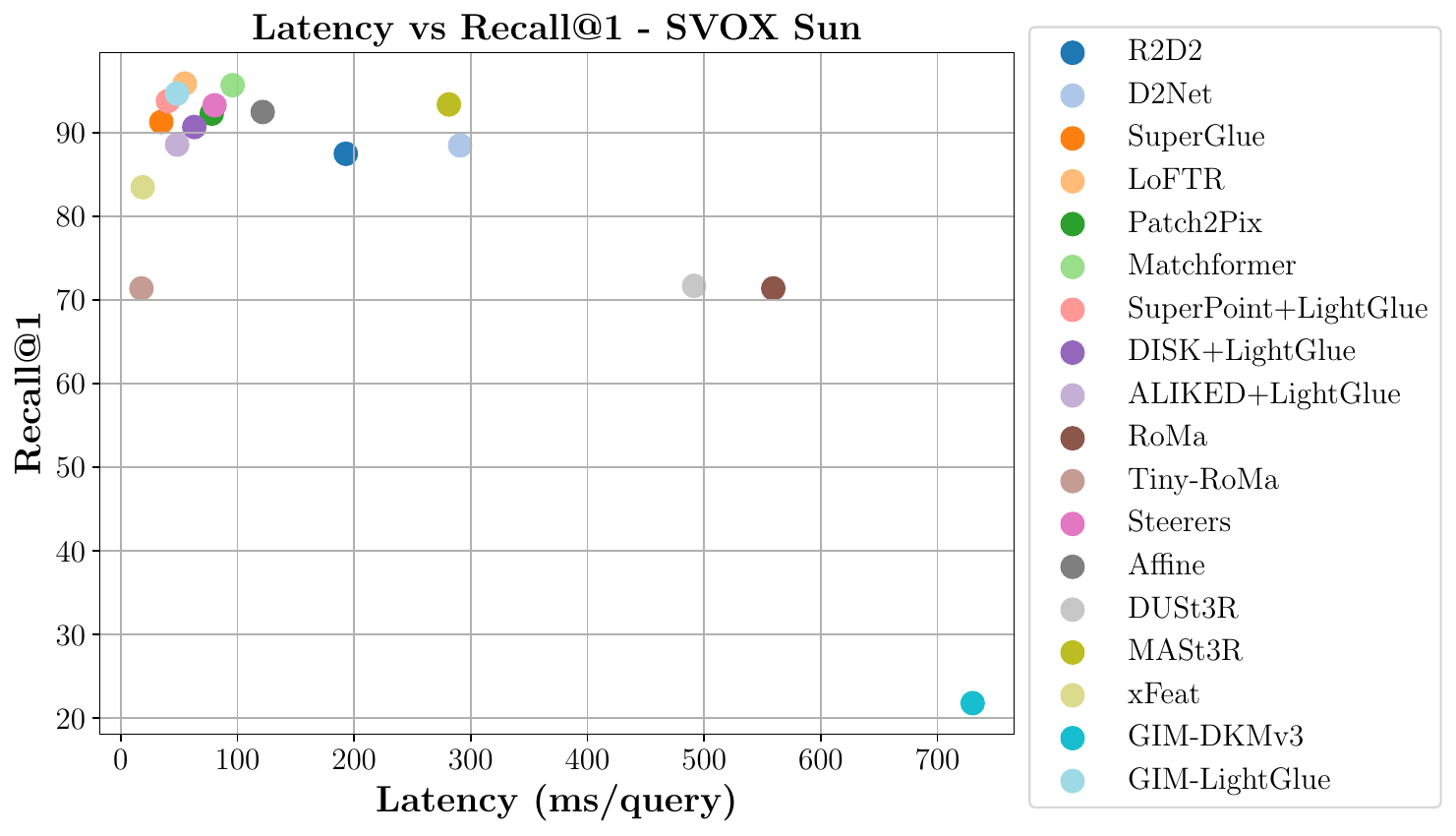}
  \label{fig:latency_svox_sun}
\end{subfigure}%
\hfill\begin{subfigure}{0.49\linewidth}
  \centering
  \includegraphics[width=0.7\linewidth]{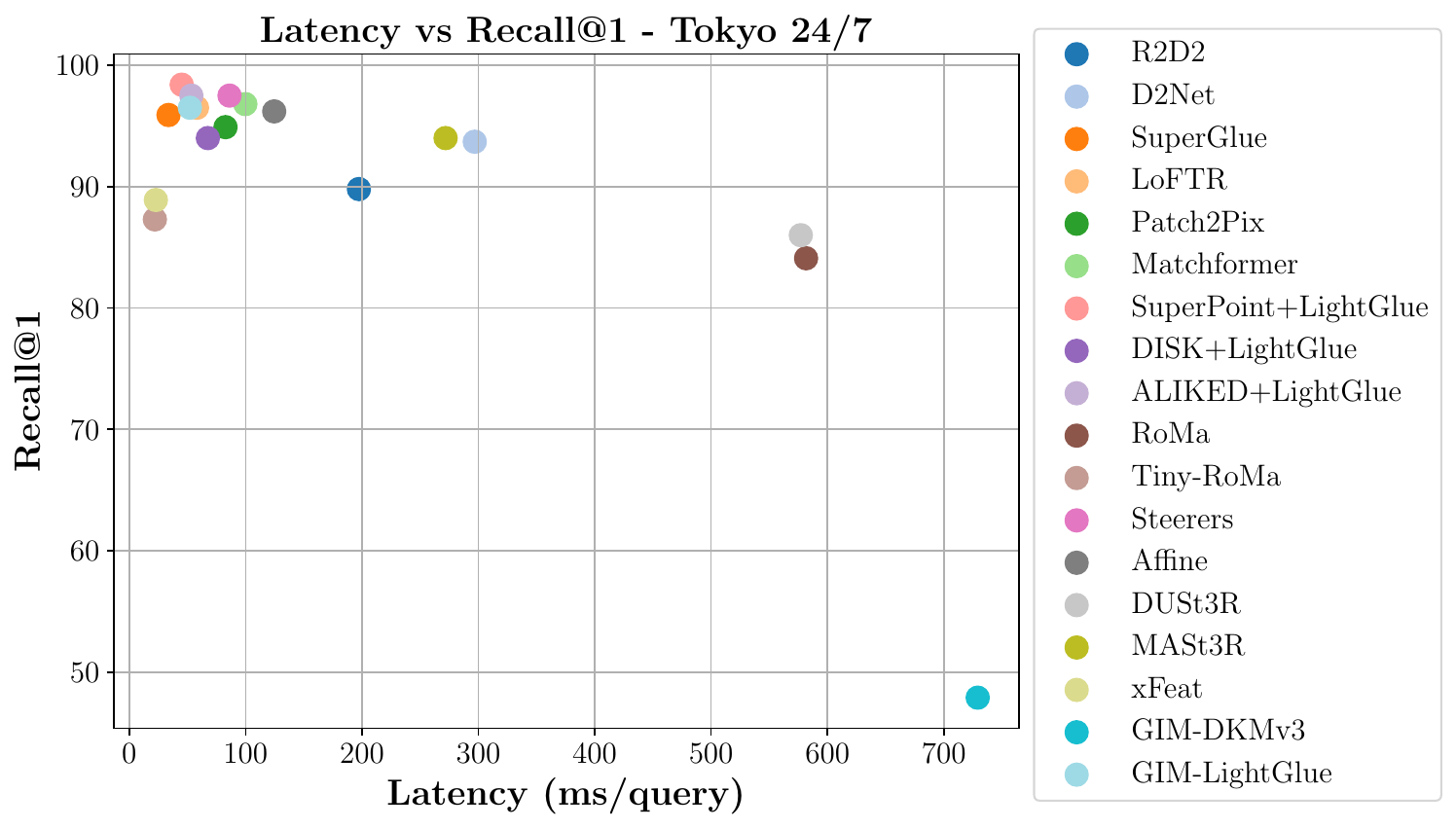}
  \label{fig:latency_tokyo247}
\end{subfigure}
\caption{\textbf{Plot displaying the Recall@1 after re-ranking and mean latency for different methods.} For each dataset, we compute the Recall@1 after re-ranking and the mean latency, which is the average time to process each query. The shortlist of candidates for the Recall@1 is obtained with MegaLoc and distance threshold fixed at 25 meters.}
\label{fig:all_latencies}
\end{figure*}

In this supplementary, we show:
\begin{itemize}
    \item Additional qualitative examples of failure cases with high number of inliers in \cref{fig:matches_remaining_datasets_supp};
    \item In \cref{fig:all_latencies}, the comprehensive set of plots showing the Recall@1 after re-ranking against the average time to process each query over each dataset;
    \item Uncertainty estimation results (in terms of AUCPR) applied on CliqueMining~\cite{Izquierdo2024-clique} predictions (the previous SOTA, before MegaLoc~\cite{Berton_2025_megaloc}), and the re-ranking performance.
    \item PR curves on the most challenging datasets for CliqueMining;
\end{itemize}

\subsection{Results with CliqueMining}
In \cref{tab:auprc_cliquemining_1_decimal_25m} we report the uncertainty estimation benchmark starting from CliqueMining predictions, and in  \cref{tab:reranking_results_clique_mining_2_recall} the re-ranking performances of several image matching methods.
While CliqueMining in general attains lower performances w.r.t. MegaLoc, these two tables confirm the findings in our main paper. Essentially, higher performances in pure retrieval diminish the usefulness of a matching step. In turn, in scenarios in which the retrieval method struggles, the number of inliers provides a reliable confidence estimate, with AUC scores of 77.1 and 85.7 for the best performing ones on SF-XL Night and Occlusion, respectively.
As for MegaLoc, on the more challenging datasets of Baidu, SF-XL Night and Occlusion, the re-ranking step remains beneficial in terms of R@1.

\subsection{PR curves for CliqueMining}
\begin{figure*}
\centering
\begin{subfigure}{0.33\textwidth}
  \centering
  \includegraphics[width=\linewidth]{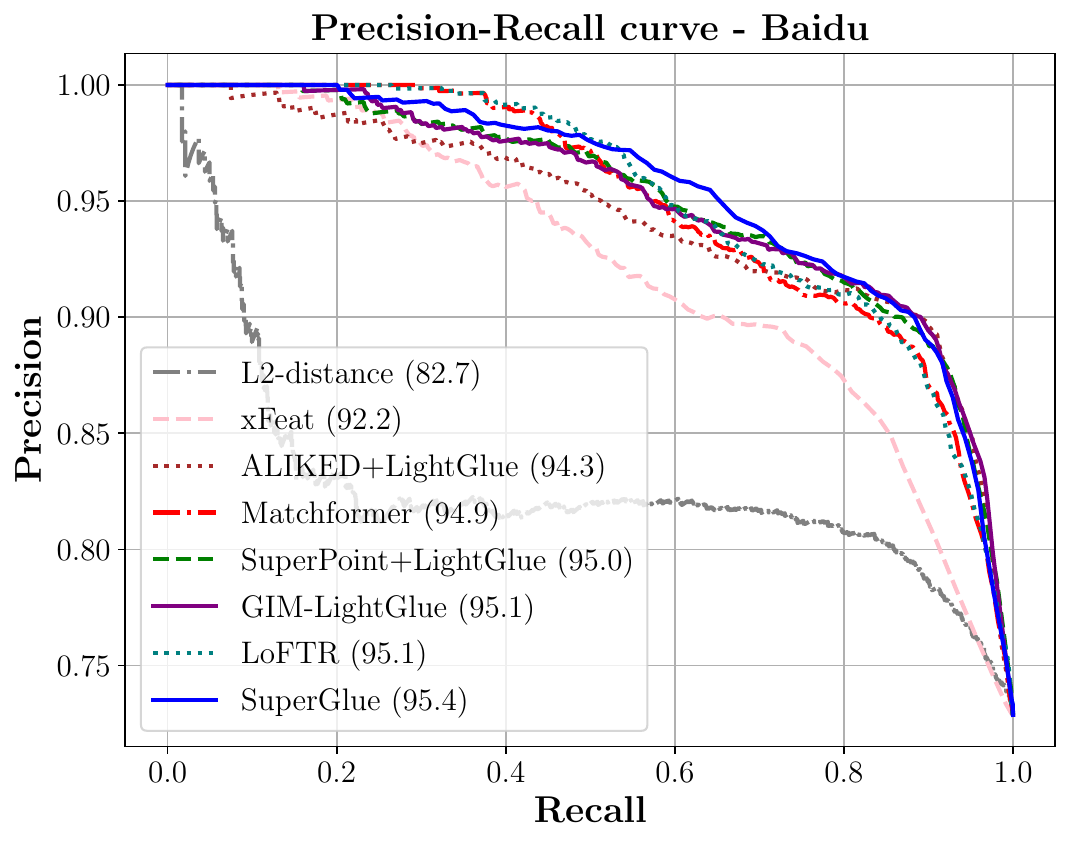}
  \label{fig:pr_curve_baidu_clique_mining}
\end{subfigure}%
\begin{subfigure}{0.33\textwidth}
  \centering
  \includegraphics[width=\linewidth]{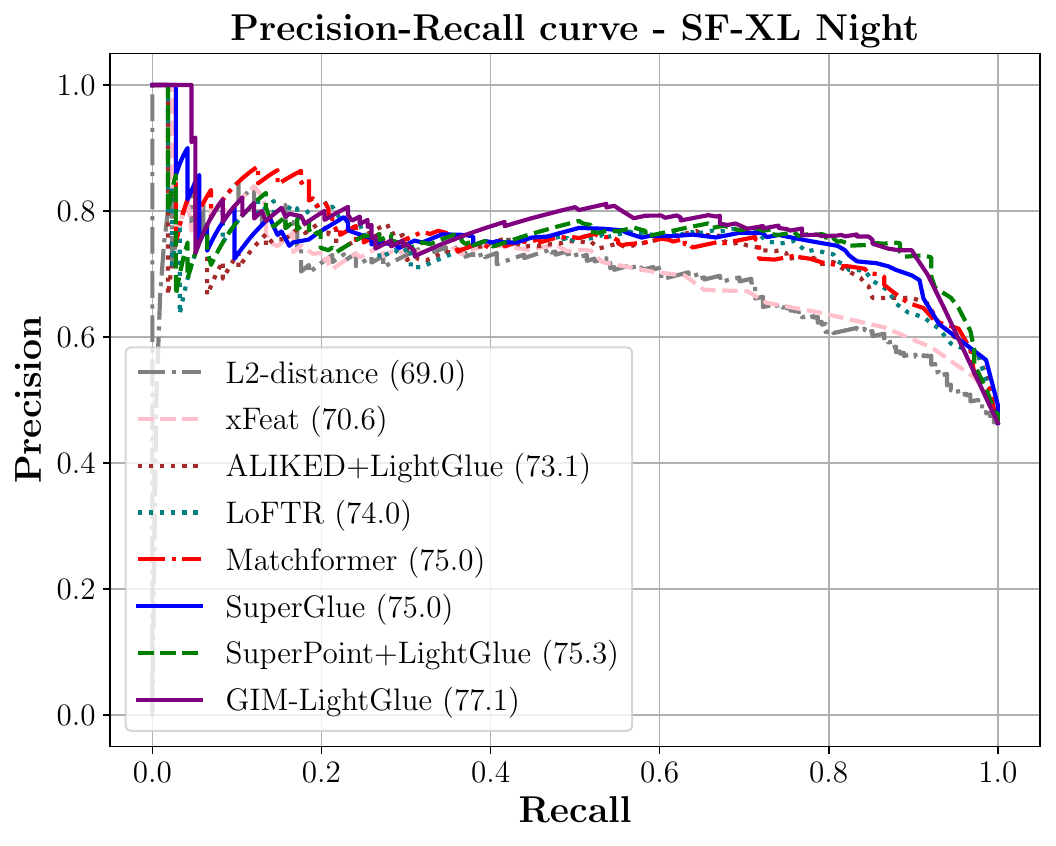}
  \label{fig:pr_curve_sf_xl_night_clique_mining}
\end{subfigure}
\begin{subfigure}{0.33\textwidth}
  \centering
  \includegraphics[width=\linewidth]{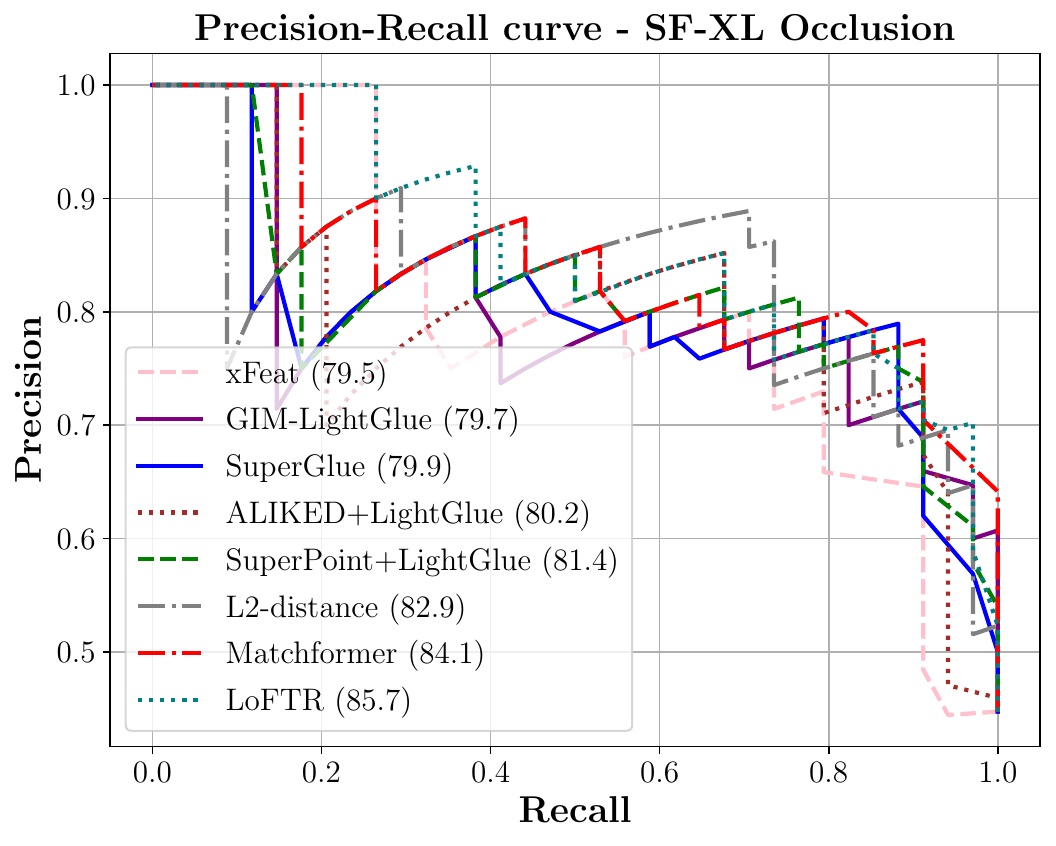}
  \label{fig:pr_curve_sf_xl_clique_mining}
\end{subfigure}
\vspace{-5mm}
\caption{\textbf{Precision-Recall curves}, computed for the top-7 image matching methods on Baidu, SF-XL Night, and SF-XL Occlusion, together with the L2 distance in embedding space. Distance threshold is fixed at 25 meters.}
\label{fig:precision_recall_curves_clique_mining}
\end{figure*}

In \cref{fig:precision_recall_curves_clique_mining} we report PR curves for the most challenging datasets of Baidu, SF-XL Night and Occlusion. We show how in these scenarios there is still room for improvement upon CliqueMining predictions, and how image matching methods outperform existing baselines for uncertainty prediction, attaining higher AUC-PR scores.

\begin{table*}
\begin{center}
\begin{adjustbox}{width=0.75\linewidth}
\centering
\setlength{\tabcolsep}{2pt}
\begin{tabular}{lcccccccccc|c}
\toprule
\multirow{2}{*}{Method} & \multirow{2}{*}{Baidu} & MSLS & \multirow{2}{*}{Pitts30k} & SF-XL & SF-XL & SF-XL & SF-XL & SVOX & SVOX & Tokyo & \multirow{2}{*}{Average} \\
& & Val & & Night & Occlusion & test V1 & test V2 & Night & Sun & 24/7 & \\
\hline
L2-distance & 82.7 & 97.2 & \underline{98.7} & 69.0 & \underline{82.9} & 98.0 & 97.9 & 98.6 & \underline{99.7} & \underline{99.9} & 92.5 \\
PA-Score & 85.1 & 96.7 & 98.6 & 63.8 & 70.7 & 96.8 & 97.9 & 98.7 & 99.5 & 99.4 & 90.7 \\
SUE & \underline{88.8} & \underline{97.3} & 98.5 & \underline{73.0} & 80.4 & \underline{98.2} & \textbf{98.6} & \underline{99.2} & 99.6 & 99.7 & \underline{93.3} \\
Random & 73.1 & 91.5 & 92.3 & 42.4 & 41.4 & 83.6 & 93.1 & 95.0 & 97.8 & 97.2 & 80.7 \\
\hline
R2D2 (NeurIPS '19) & 93.7 & 96.6 & 98.3 & 65.0 & 76.3 & 98.6 & 97.9 & 99.3 & 99.5 & 99.7 & 92.5 \\
D2Net (CVPR '19) & 95.1 & 96.4 & 98.2 & 74.1 & 77.3 & 99.4 & 97.9 & 99.5 & 99.6 & \textbf{100.0} & 93.7 \\
SuperGlue (CVPR '20) & \textbf{95.4} & 97.4 & 98.7 & 75.0 & 79.9 & 99.6 & 97.8 & 99.6 & 99.4 & \textbf{100.0} & 94.3 \\
LoFTR (CVPR '21) & 95.1 & 97.1 & 98.7 & 74.0 & \textbf{85.7} & 99.4 & \underline{98.2} & 99.6 & 99.5 & \textbf{100.0} & \textbf{94.7} \\
Patch2Pix (CVPR '21) & 94.0 & 97.0 & 98.5 & 73.0 & 78.2 & 99.3 & 97.7 & 99.6 & 99.4 & 99.9 & 93.6 \\
Matchformer (ACCV '22) & 94.9 & 97.1 & 98.8 & 75.0 & 84.1 & 99.5 & 98.1 & \textbf{99.7} & 99.4 & \textbf{100.0} & \textbf{94.7} \\
SuperPoint+LightGlue (ICCV '23) & 95.0 & 97.6 & 98.8 & 75.3 & 81.4 & 99.6 & 97.9 & \textbf{99.7} & 99.3 & \textbf{100.0} & 94.5 \\
DISK+LightGlue (ICCV '23) & 91.8 & 97.2 & 98.9 & 72.2 & 81.2 & 99.4 & 97.7 & 99.4 & 99.3 & \textbf{100.0} & 93.7 \\
ALIKED+LightGlue (ICCV '23) & 94.3 & \textbf{97.8} & 99.0 & 73.1 & 80.2 & 99.5 & 98.1 & \textbf{99.7} & 99.6 & \textbf{100.0} & 94.1 \\
RoMa (CVPR '24) & 89.7 & 96.4 & 96.7 & 70.4 & 61.5 & 97.6 & 96.1 & \textbf{99.7} & 99.5 & \textbf{100.0} & 90.8 \\
Tiny-RoMa (CVPR '24) & 94.8 & 96.7 & 98.8 & 69.6 & 77.7 & 98.8 & 97.4 & 98.7 & 99.3 & 99.8 & 93.2 \\
Steerers (CVPR '24) & 93.9 & 97.1 & \textbf{99.1} & 72.8 & 76.8 & 99.5 & 98.1 & \textbf{99.7} & 99.2 & \textbf{100.0} & 93.6 \\
Affine Steerers (ECCV '24) & 93.4 & 97.1 & 98.7 & 69.7 & 74.0 & 99.4 & 98.1 & 99.5 & 99.0 & 99.9 & 92.9 \\
DUSt3R (CVPR '24) & 89.5 & 97.4 & 98.6 & 69.6 & 60.8 & 95.1 & 96.4 & 98.7 & 99.5 & 99.4 & 90.5 \\
MASt3R (ECCV '24) & 90.4 & 97.0 & \textbf{99.1} & 72.7 & 80.1 & 99.6 & 97.0 & \textbf{99.7} & \textbf{99.9} & \textbf{100.0} & 93.5 \\
xFeat (CVPR '24) & 92.1 & 96.9 & 98.6 & 70.6 & 79.5 & 98.9 & 97.1 & 99.2 & 99.3 & 99.8 & 93.2 \\
GIM-DKMv3 (ICLR '24) & 84.4 & 93.0 & 95.0 & 67.1 & 74.5 & 92.3 & 94.8 & 98.3 & 99.3 & 99.9 & 89.9 \\
GIM-LightGlue (ICLR '24) & 95.1 & 97.5 & 98.8 & \textbf{77.1} & 79.7 & \textbf{99.7} & 97.8 & \textbf{99.7} & 99.5 & \textbf{100.0} & 94.5 \\
\bottomrule
\end{tabular}
\end{adjustbox}
\end{center}
\vspace{-3mm}
\caption{\textbf{The AUPRC of all the baselines and image matching methods}, split according to group type. The shorlist of candidates is obtained with CliqueMining. Distance threshold is fixed at 25 meters. Best
overall results on each dataset are in \textbf{bold}, best results for each group are \underline{underlined}.}
\label{tab:auprc_cliquemining_1_decimal_25m}
\end{table*}
\begin{table*}
\begin{adjustbox}{width=\linewidth}
\centering
\setlength{\tabcolsep}{2pt}
\begin{tabular}{lcccccccccccccccccccccccccccccc@{}|@{}ccc}
\toprule
\multirow{3}{*}{{\begin{tabular}[c]{@{}c@{}}Method\end{tabular}}} &
\multicolumn{2}{c}{Baidu} & & 
\multicolumn{2}{c}{MSLS Val} & & 
\multicolumn{2}{c}{Pitts30k} & & 
\multicolumn{2}{c}{SF-XL Night} & & 
\multicolumn{2}{c}{SF-XL Occlusion} & &
\multicolumn{2}{c}{SF-XL test V1} & &
\multicolumn{2}{c}{SF-XL test V2} & &
\multicolumn{2}{c}{SVOX Night} & &
\multicolumn{2}{c}{SVOX Sun} & &
\multicolumn{2}{c}{Tokyo 24/7} & & &
\multicolumn{2}{c}{Average} \\
& 
\multicolumn{2}{c}{R@20 = 96.9} & & 
\multicolumn{2}{c}{R@20 = 96.6} & & 
\multicolumn{2}{c}{R@20 = 98.6} & & 
\multicolumn{2}{c}{R@20 = 64.8} & & 
\multicolumn{2}{c}{R@20 = 65.8} & & 
\multicolumn{2}{c}{R@20 = 94.0} & & 
\multicolumn{2}{c}{R@20 = 98.5} & & 
\multicolumn{2}{c}{R@20 = 99.9} & & 
\multicolumn{2}{c}{R@20 = 99.5} & & 
\multicolumn{2}{c}{R@20 = 98.4} & & &
\multicolumn{2}{c}{R@20 = 91.3} \\
\cline{2-3} \cline{5-6} \cline{8-9} \cline{11-12} \cline{14-15} \cline{17-18} \cline{20-21} \cline{23-24} \cline{26-27} \cline{29-30} \cline{33-34}
& R@1 & R@10 & & R@1 & R@10 & & R@1 & R@10 & & R@1 & R@10 & & R@1 & R@10 & & R@1 & R@10 & & R@1 & R@10 & & R@1  & R@10 & & R@1  & R@10 & & R@1 & R@10 & & & R@1 & R@10 \\
\hline

- & 72.9 & 92.7 && \textbf{91.6} & \textbf{95.9} && \textbf{92.6} & 97.8 && 46.1 & 60.9 && 44.7 & \underline{64.5} && 85.5 & 92.6 && 94.5 & \underline{98.3} && \textbf{95.5} & \underline{99.6} && \textbf{98.2} & \underline{99.4} && 96.8 & 97.8 &&& 81.8 & 90.0 \\

\hline

R2D2 (NeurIPS '19) & 83.2 & 94.9 && 82.1 & 94.7 && 88.6 & 97.6 && 36.9 & 59.2 && 46.1 & 60.5 && 85.7 & 92.3 && 94.1 & 98.2 && 80.8 & 97.3 && 91.7 & 98.8 && 91.1 & \underline{98.1} &&& 78.0 & 89.2 \\
D2Net (CVPR '19) & 83.8 & 95.1 && 83.4 & 95.2 && 89.4 & 97.7 && 44.8 & 62.4 && \textbf{53.9} & 59.2 && 88.8 & 93.4 && 95.0 & \textbf{98.5} && 86.3 & 98.2 && 93.1 & 98.9 && 95.6 & 97.8 &&& 81.4 & 89.6 \\
SuperGlue (CVPR '20) & 84.9 & 95.0 && 87.7 & \underline{95.8} && 89.5 & \underline{98.0} && 51.3 & \textbf{63.5} && 48.7 & 61.8 && \textbf{91.3} & \underline{93.9} && 94.5 & \underline{98.3} && 91.5 & 98.8 && 93.8 & \textbf{99.5} && \underline{97.5} & 97.8 &&& 83.1 & 90.2 \\
LoFTR (CVPR '21) & 85.9 & 95.2 && 88.1 & 95.6 && 90.0 & 97.6 && 51.3 & \underline{62.9} && 50.0 & \underline{64.5} && \underline{91.2} & 93.5 && \underline{95.5} & \textbf{98.5} && 93.7 & 99.4 && \underline{96.8} & \underline{99.4} && \underline{97.5} & \underline{98.1} &&& \underline{84.0} & \underline{90.5} \\
Patch2Pix (CVPR '21) & 83.6 & 95.2 && 84.8 & 95.1 && 88.8 & 97.8 && 41.2 & 59.9 && 51.3 & 61.8 && 88.8 & 93.1 && 94.8 & \underline{98.3} && 79.8 & 99.3 && 95.3 & \underline{99.4} && 96.8 & \underline{98.1} &&& 80.5 & 89.8 \\
Matchformer (ACCV '22) & 85.8 & 95.1 && 86.5 & 95.2 && 90.7 & 97.9 && 50.6 & 61.6 && 50.0 & 63.2 && 91.0 & 93.7 && \textbf{95.7} & \textbf{98.5} && 94.5 & 99.1 && 96.1 & \textbf{99.5} && \textbf{97.8} & \textbf{98.4} &&& 83.9 & 90.2 \\
SuperPoint+LightGlue (ICCV '23) & \underline{86.0} & 95.2 && 87.3 & \underline{95.8} && 90.3 & \underline{98.0} && \textbf{53.0} & 62.7 && 47.4 & 63.2 && \underline{91.2} & \textbf{94.0} && 93.6 & \textbf{98.5} && 92.2 & 99.4 && 95.7 & 99.3 && 97.1 & \underline{98.1} &&& 83.4 & 90.4 \\
DISK+LightGlue (ICCV '23) & 84.4 & \underline{95.4} && 87.0 & 95.5 && 89.3 & 97.8 && 50.4 & 62.4 && 48.7 & \underline{64.5} && 89.1 & 93.6 && 92.8 & \underline{98.3} && 85.9 & 97.6 && 94.1 & 99.3 && 96.5 & \underline{98.1} &&& 81.8 & 90.2 \\
ALIKED+LightGlue (ICCV '23) & \underline{86.0} & \textbf{95.5} && \underline{89.2} & 95.7 && 89.4 & \underline{98.0} && 50.4 & \textbf{63.5} && 51.3 & \underline{64.5} && 90.3 & 93.6 && 94.5 & \underline{98.3} && 88.0 & 99.0 && 94.6 & 99.3 && 97.1 & \underline{98.1} &&& 83.1 & \underline{90.5} \\
RoMa (CVPR '24) & 81.1 & 95.0 && 82.9 & 95.2 && 87.4 & 97.6 && 42.1 & 62.0 && 47.4 & 63.2 && 79.4 & 93.5 && 93.5 & \textbf{98.5} && 85.4 & \textbf{99.8} && 87.8 & \textbf{99.5} && 93.7 & 97.8 &&& 78.1 & 90.2 \\
Tiny-RoMa (CVPR '24) & 82.4 & \underline{95.4} && 82.9 & 95.2 && 87.4 & 97.6 && 42.1 & 62.0 && 47.4 & 63.2 && 87.1 & 93.3 && 93.5 & \textbf{98.5} && 68.7 & 97.8 && 82.7 & 99.2 && 93.7 & 97.8 &&& 76.8 & 90.0 \\
Steerers (CVPR '24) & 85.7 & 95.0 && 81.4 & 94.4 && 89.7 & 97.7 && 44.8 & 61.8 && 50.0 & 61.8 && 90.3 & 93.7 && 95.0 & \textbf{98.5} && 88.5 & 99.3 && 95.3 & 99.3 && 96.5 & \textbf{98.4} &&& 81.7 & 90.0 \\
Affine Steerers (ECCV '24) & 84.4 & 95.1 && 84.2 & 95.3 && 89.1 & 97.7 && 47.4 & 62.4 && 48.7 & 59.2 && 89.5 & 93.7 && 94.5 & 98.2 && 88.5 & 99.3 && 94.3 & 99.1 && 96.8 & \underline{98.1} &&& 81.7 & 89.8 \\
DUSt3R (CVPR '24) & 77.2 & 94.7 && 75.3 & 89.5 && 83.0 & 95.6 && 35.0 & 53.9 && 36.8 & 52.6 && 77.1 & 87.0 && 76.9 & 97.7 && 67.4 & 79.1 && 77.9 & 86.2 && 88.3 & 96.8 &&& 69.5 & 83.3 \\
MASt3R (ECCV '24) & 83.4 & 95.3 && 80.5 & 95.7 && 90.9 & \textbf{98.1} && 48.1 & 62.2 && 51.3 & \textbf{65.8} && 90.3 & 93.8 && 93.5 & \underline{98.3} && \underline{95.3} & \underline{99.6} && 96.3 & \underline{99.4} && 96.8 & \textbf{98.4} &&& 82.6 & \textbf{90.7} \\
xFeat (CVPR '24) & 81.4 & 94.1 && 85.8 & 95.0 && 89.4 & 97.7 && 43.6 & 61.2 && 43.4 & 60.5 && 86.9 & 93.0 && 93.5 & \textbf{98.5} && 83.1 & 98.2 && 89.0 & 98.6 && 93.0 & 97.8 &&& 78.9 & 89.5 \\
GIM-DKMv3 (ICLR '24) & 53.0 & 95.2 && 32.1 & 92.3 && 66.2 & 97.7 && 38.2 & 62.4 && 36.8 & 63.2 && 53.8 & 93.5 && 75.3 & 98.0 && 55.2 & \textbf{99.8} && 51.8 & 99.3 && 77.8 & \underline{98.1} &&& 54.0 & 90.0 \\
GIM-LightGlue (ICLR '24) & \textbf{86.5} & \underline{95.4} && 88.6 & 95.7 && \underline{91.1} & 97.8 && \underline{51.5} & 61.4 && \underline{52.6} & \underline{64.5} && 91.1 & 93.8 && 94.3 & \underline{98.3} && 93.1 & 97.8 && \underline{96.8} & 98.9 && 96.5 & \underline{98.1} &&& \textbf{84.2} & 90.2 \\
\bottomrule
\end{tabular}
\end{adjustbox}
\vspace{-3mm}
\caption{\textbf{Recalls before and after applying re-ranking.} Recalls are computed by setting the distance threshold to 25 meters.
The shortlist of candidates to be re-ranked is obtained with CliqueMining, and the results with such shortlist are shown in the first row.
Re-ranking has been applied to the first 20 candidates (\ie $K=20$). Next to each dataset's name, we show the R@20, which in practice sets the upper bound of the maximum recalls achievable after re-ranking. Best results are in \textbf{bold}, second best are \underline{underlined}.
}
\label{tab:reranking_results_clique_mining_2_recall}
\end{table*}

\end{document}